\documentclass[letterpaper, 10 pt, conference]{ieeeconf}  

\IEEEoverridecommandlockouts                              

\usepackage{algorithm}
\usepackage{algorithmic}
\usepackage{url}
\usepackage{amsmath}
\usepackage{cite}
\usepackage[dvips]{graphicx}
\usepackage{epsfig}
\usepackage{epic,eepic,color}
\usepackage{times}
\usepackage{mathptmx}
\usepackage{multirow}
\usepackage{stkernel}

\usepackage{cases}

\usepackage{times}
\usepackage{multirow}

\definecolor{red}{rgb}{1,0,0}
\definecolor{green}{rgb}{0,1,0}
\definecolor{blue}{rgb}{0,0,1}
\definecolor{violet}{rgb}{1,0,1}
\definecolor{cyan}{cmyk}{1,0,0,0}
\definecolor{magenta}{cmyk}{0,1,0,0}
\definecolor{yellow}{cmyk}{0,0,1,0}
\definecolor{white}{rgb}{1,1,1}
\definecolor{black}{rgb}{0,0,0}

\definecolor{white}{rgb}{1,1,1}

\usepackage{jumoline}

\newcommand{\CommentOut}[1]{}

\newcommand{\FIG}[3]{
\begin{minipage}[b]{#1cm}
\begin{center}
\includegraphics[width=#1cm]{#2}
{\scriptsize #3}
\end{center}
\end{minipage}
}

\newcommand{\FIGR}[3]{
\begin{minipage}[b]{#1cm}
\begin{center}
\includegraphics[angle=-90,clip,width=#1cm]{#2}\vspace*{1mm}
\\
{\scriptsize #3}
\vspace*{1mm}
\end{center}
\end{minipage}
}

\newcommand{\FIGpng}[5]{
\begin{minipage}[b]{#1cm}
\begin{center}
\includegraphics[bb=0 0 #4 #5, clip, width=#1cm]{#2}\vspace*{-1mm}\\
{\scriptsize #3}
\vspace*{1mm}
\end{center}
\end{minipage}
}

\newcommand{\CO}[1]{}

\CO{
\renewcommand{\FIGpng}[5]{\ \ }
\renewcommand{\FIGR}[3]{\ \ }
}

\begin{document}

\author{Tanaka Kanji
\thanks{Our work has been supported in part by JSPS KAKENHI Grant-in-Aid for Young Scientists (B) 23700229, and for Scientific Research (C) 26330297.}
\thanks{K. Tanaka is with Graduate School of Engineering, University of Fukui, Japan.
{\tt\small tnkknj@u-fukui.ac.jp}}%
\vspace*{-5mm}}

\thispagestyle{empty}
\pagestyle{empty}

\newcommand{\figP}{
\begin{figure*}[t]
\begin{center}
\begin{center}
\FIGR{6}{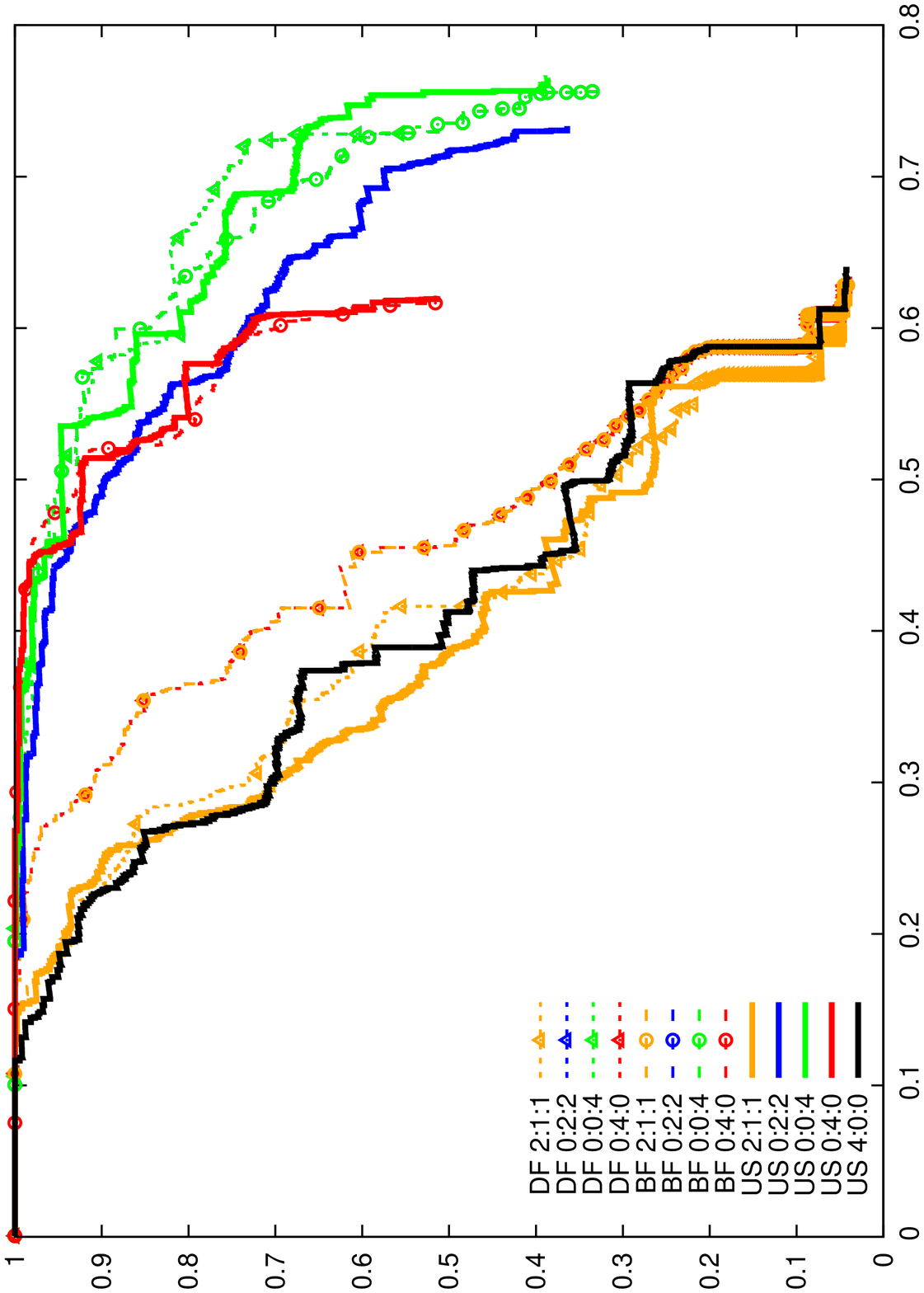}{}\hspace*{-5mm}%
\FIGR{6}{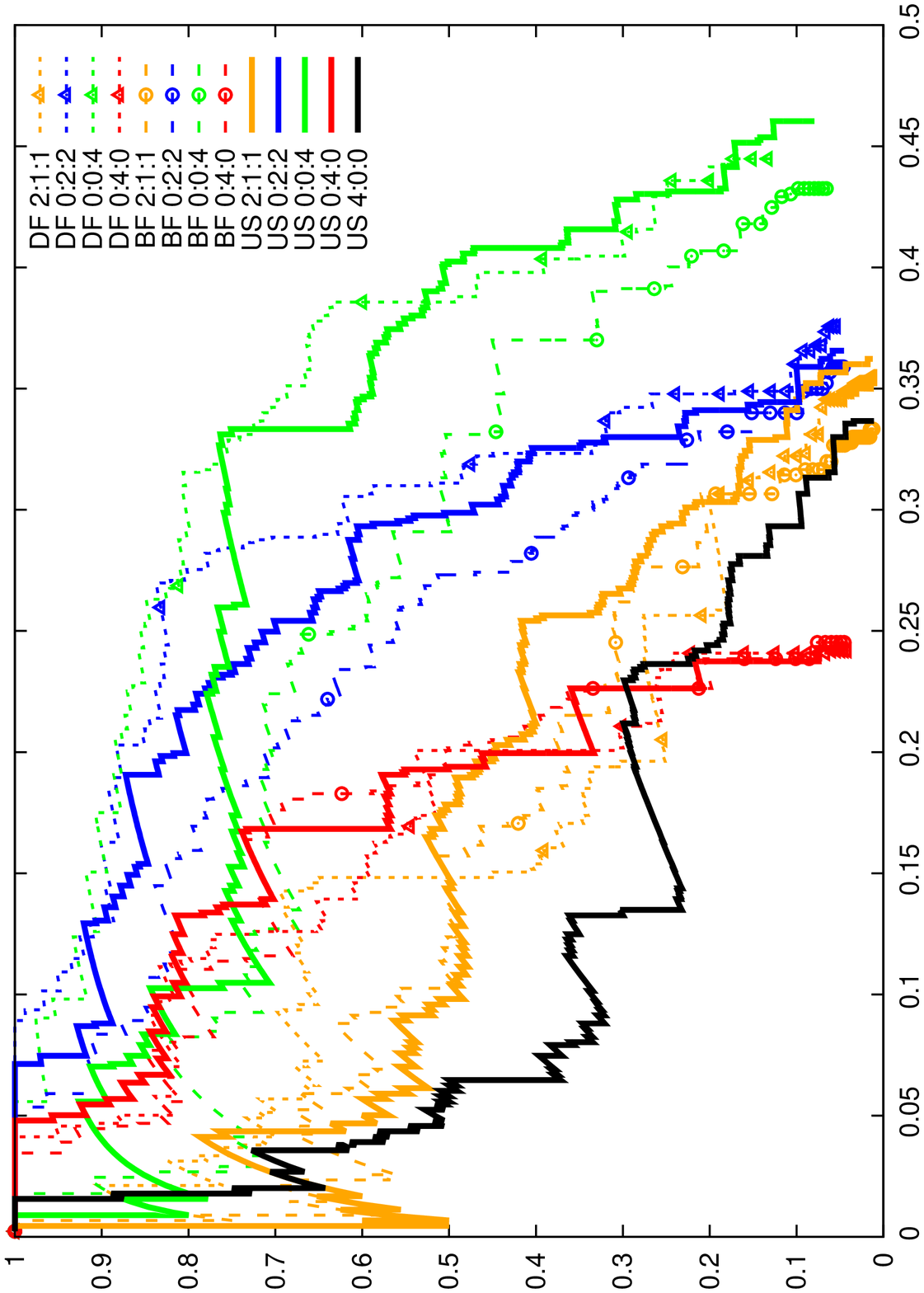}{}\hspace*{-5mm}%
\FIGR{6}{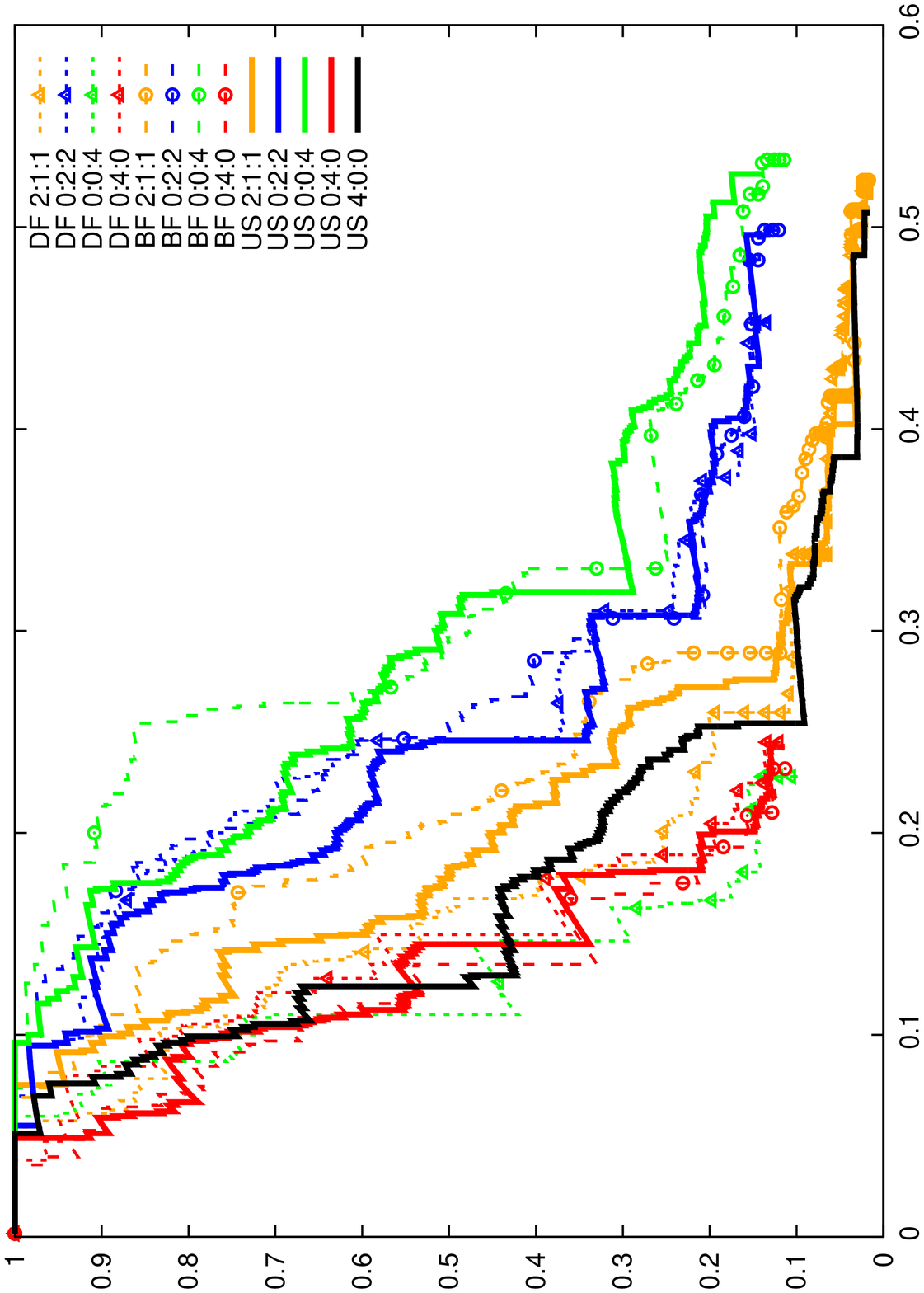}{}\vspace*{-5mm}\\
{\scriptsize a}\\
\FIGR{6}{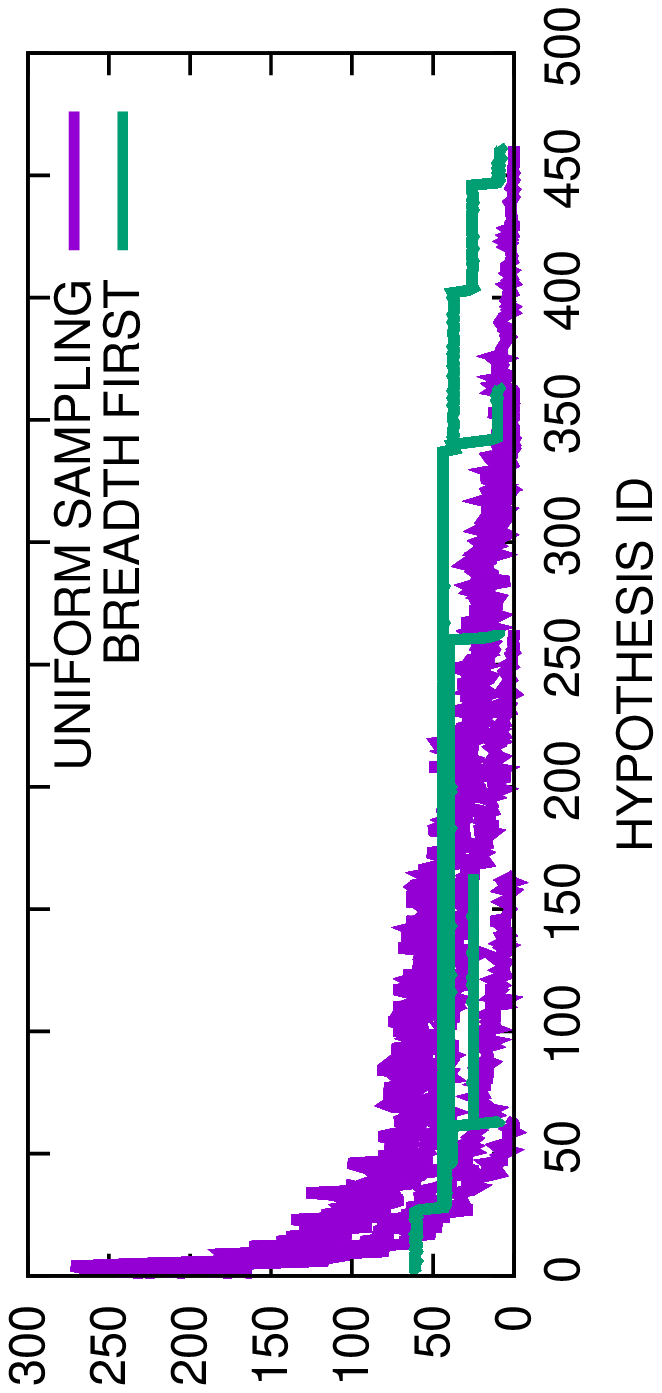}{}\hspace*{-5mm}%
\FIGR{6}{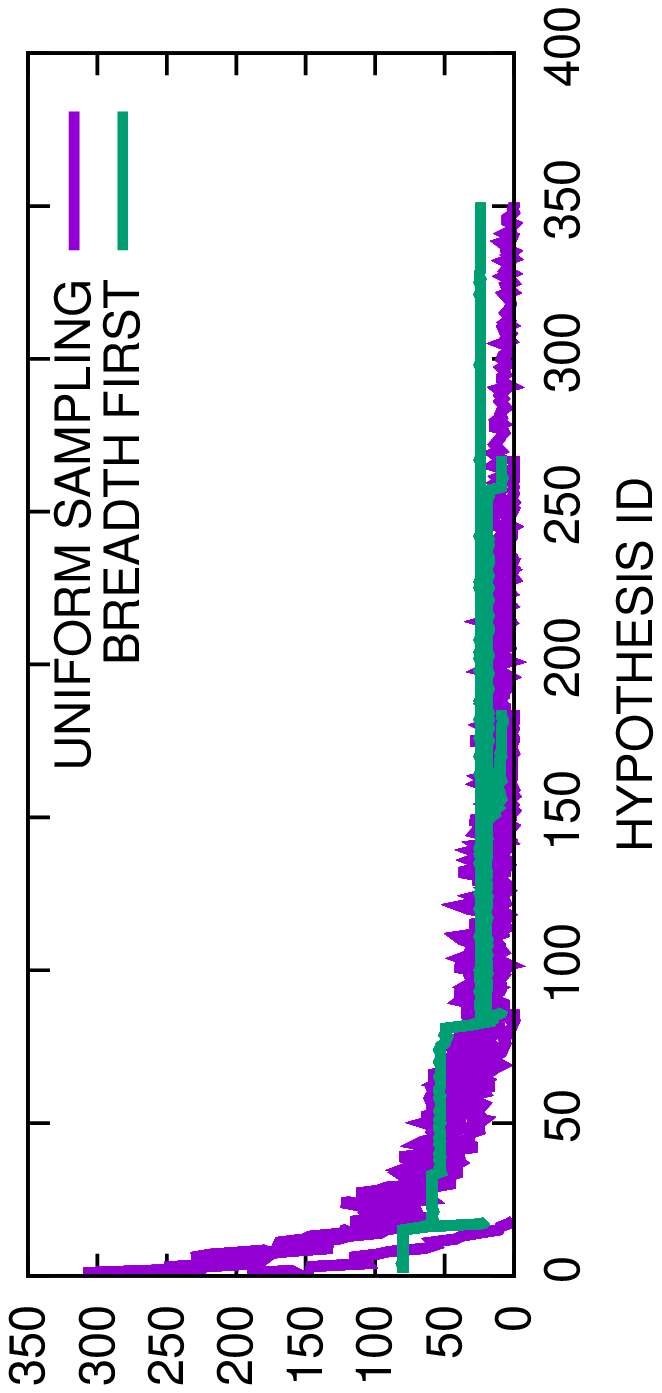}{}\hspace*{-5mm}%
\FIGR{6}{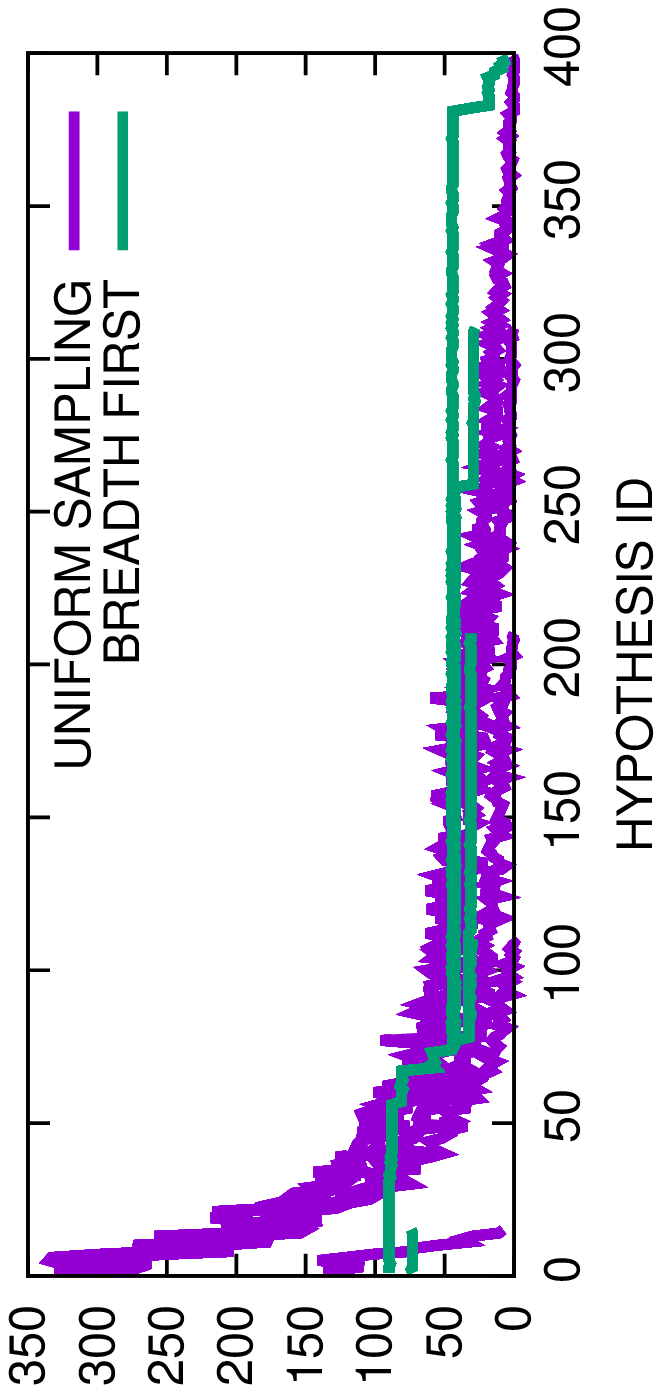}{}\vspace*{-5mm}\\
{\scriptsize b}\\
\FIGR{6}{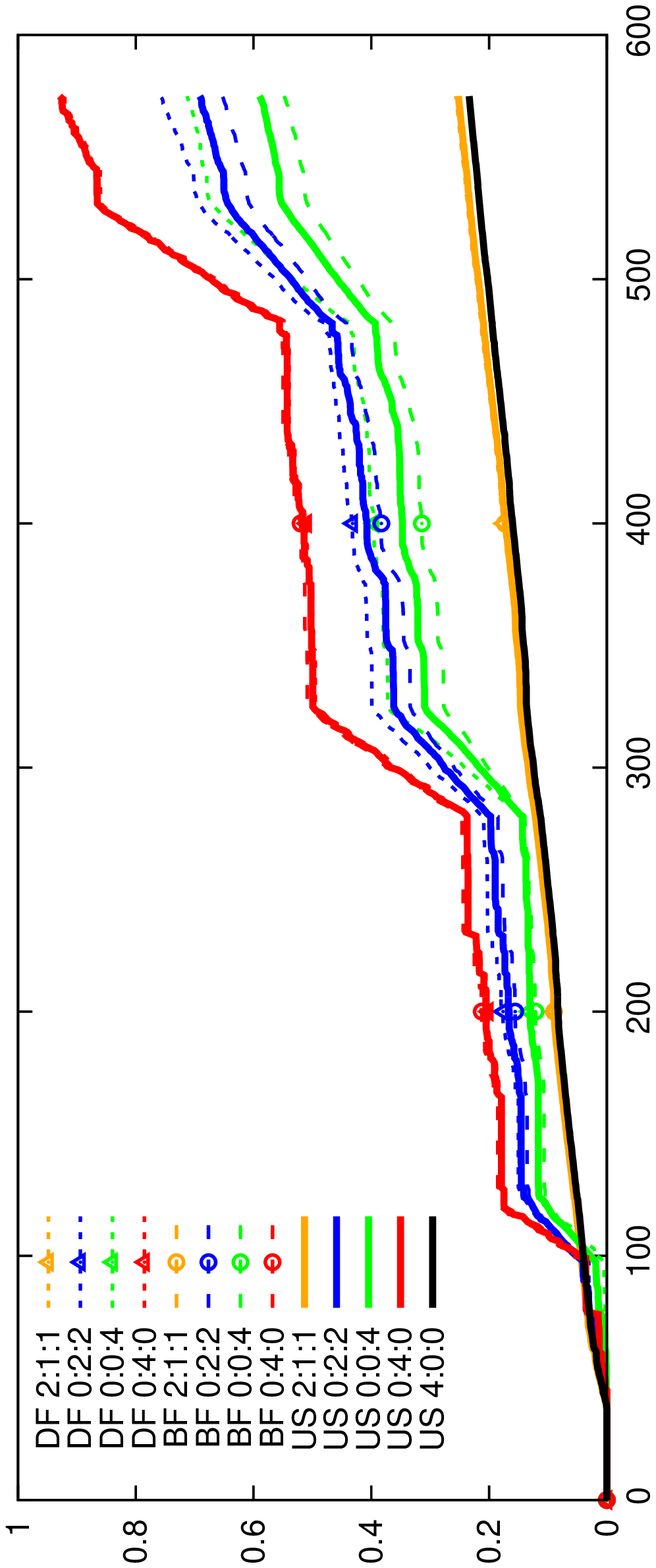}{}\hspace*{-5mm}%
\FIGR{6}{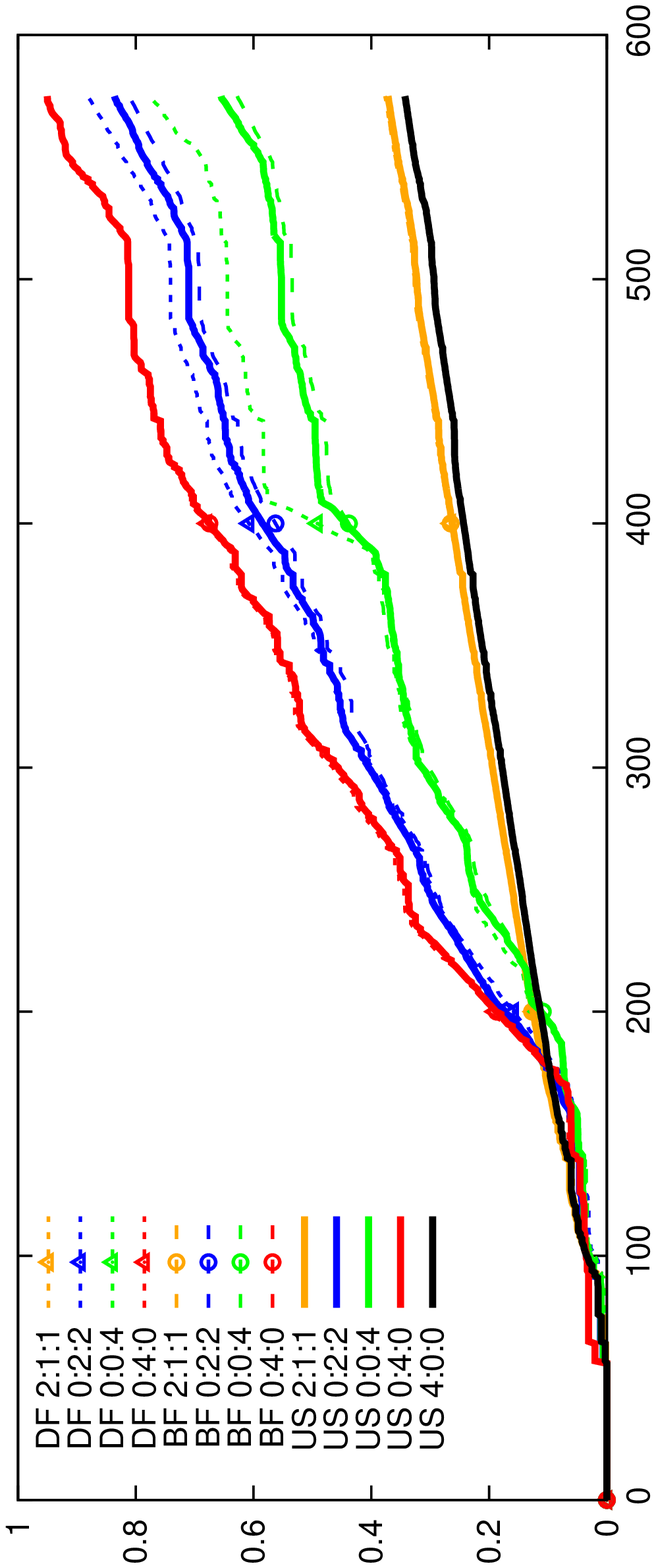}{}\hspace*{-5mm}%
\FIGR{6}{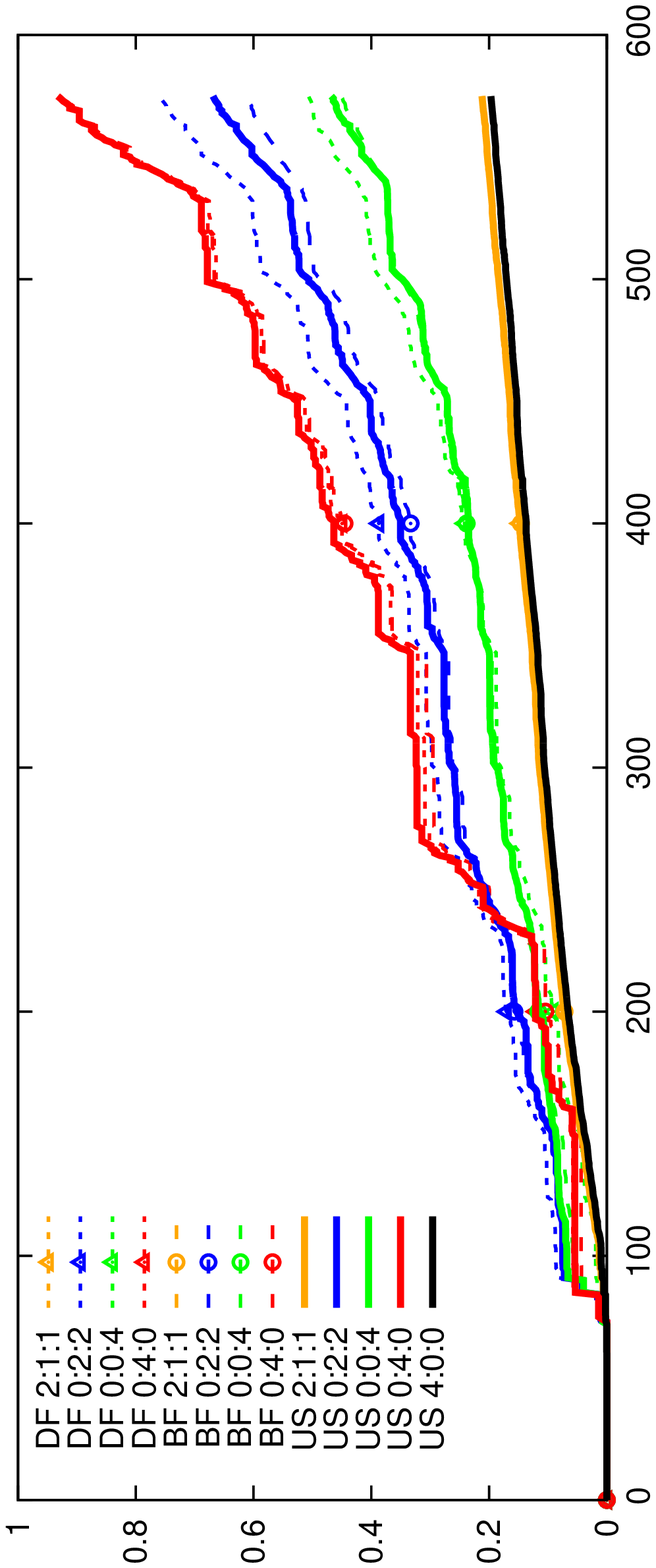}{}\vspace*{-5mm}\\
{\scriptsize c}\\
\FIGR{6}{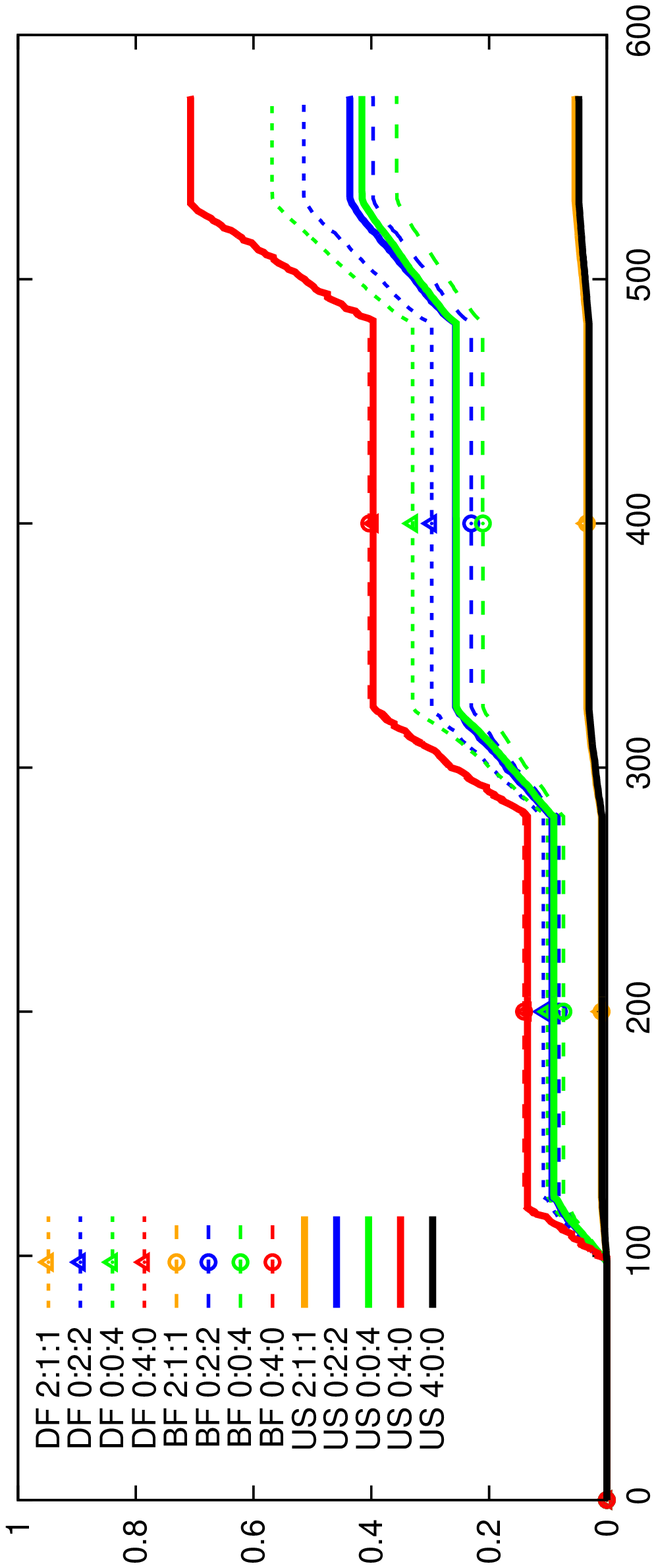}{}\hspace*{-5mm}%
\FIGR{6}{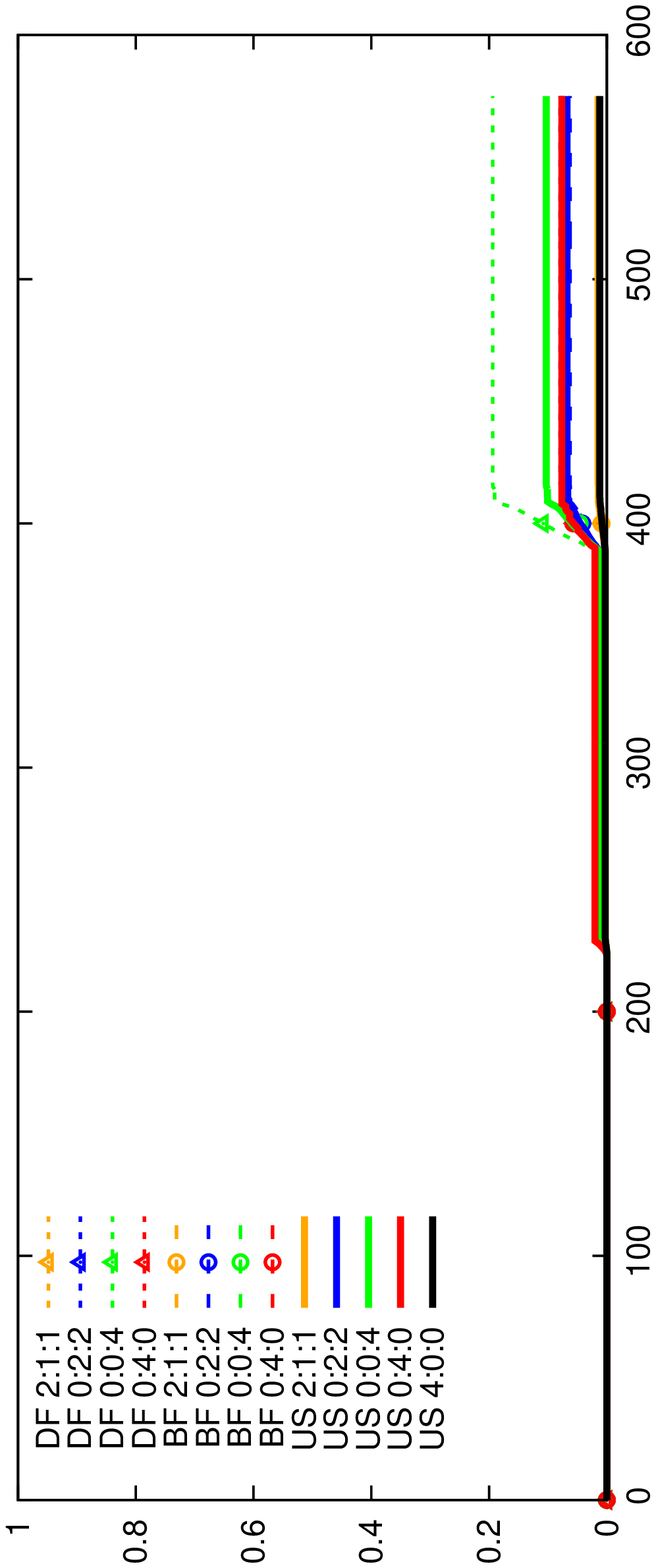}{}\hspace*{-5mm}%
\FIGR{6}{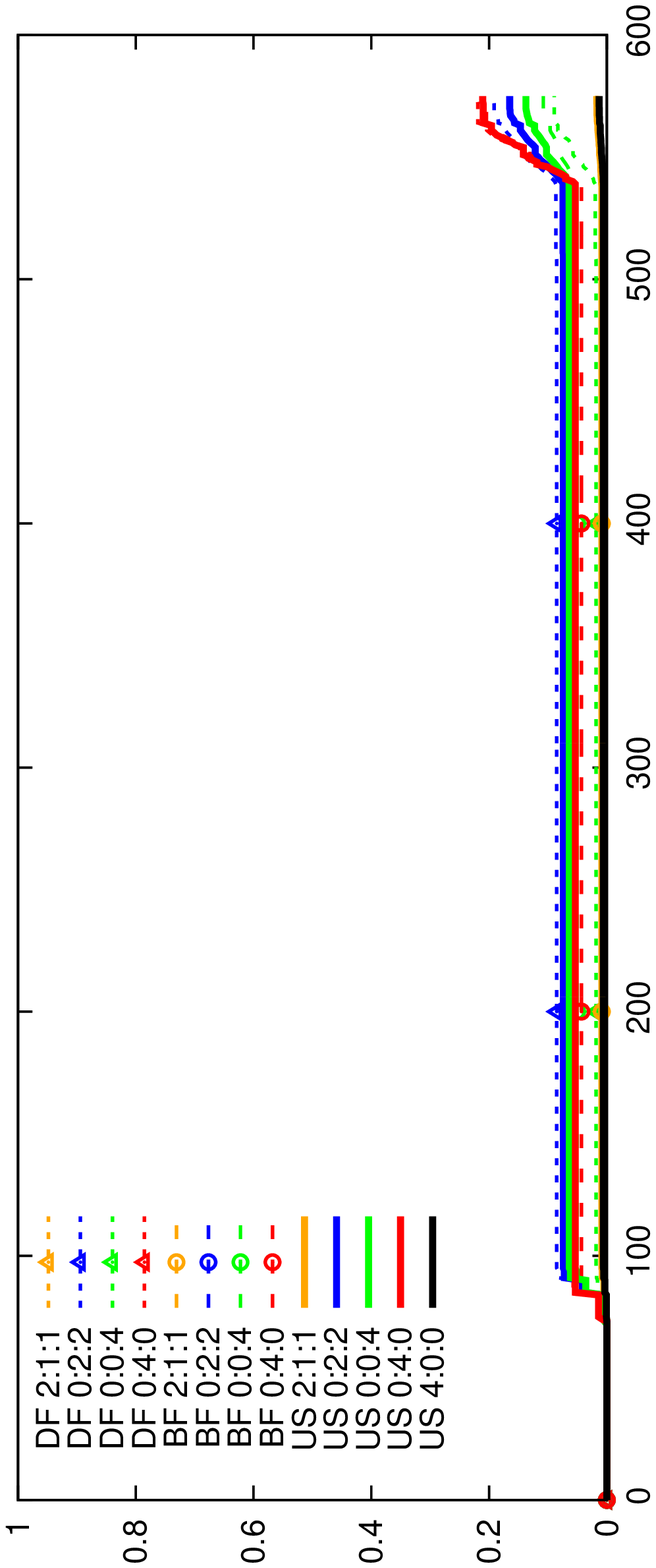}{}\vspace*{-5mm}\\
{\scriptsize d}\\
\end{center}
\caption{Performance results. From left to right, the results for three different experiments, \#1, \#2, and \#3, are presented. From top to bottom, each panel presents (a) the precision-recall curve, (b) a comparison of total number of individual hypotheses being sampled between uniform sampling and breadth first strategies, (c) ratio of loop closure constraints that are guided by individual strategies and also verified as matched by RANSAC (horizontal axis: time window ID), and (d) ratio of loop closure constraints that are guided by individual strategies, verified as matched by RANSAC, and are correct constraints with respect to the ground-truth (horizontal axis: time window ID). }\label{fig:P}
\end{center}
\end{figure*}
}

\newcommand{\figS}{
\begin{figure}[b]
\begin{center}
\begin{center}
\FIG{8}{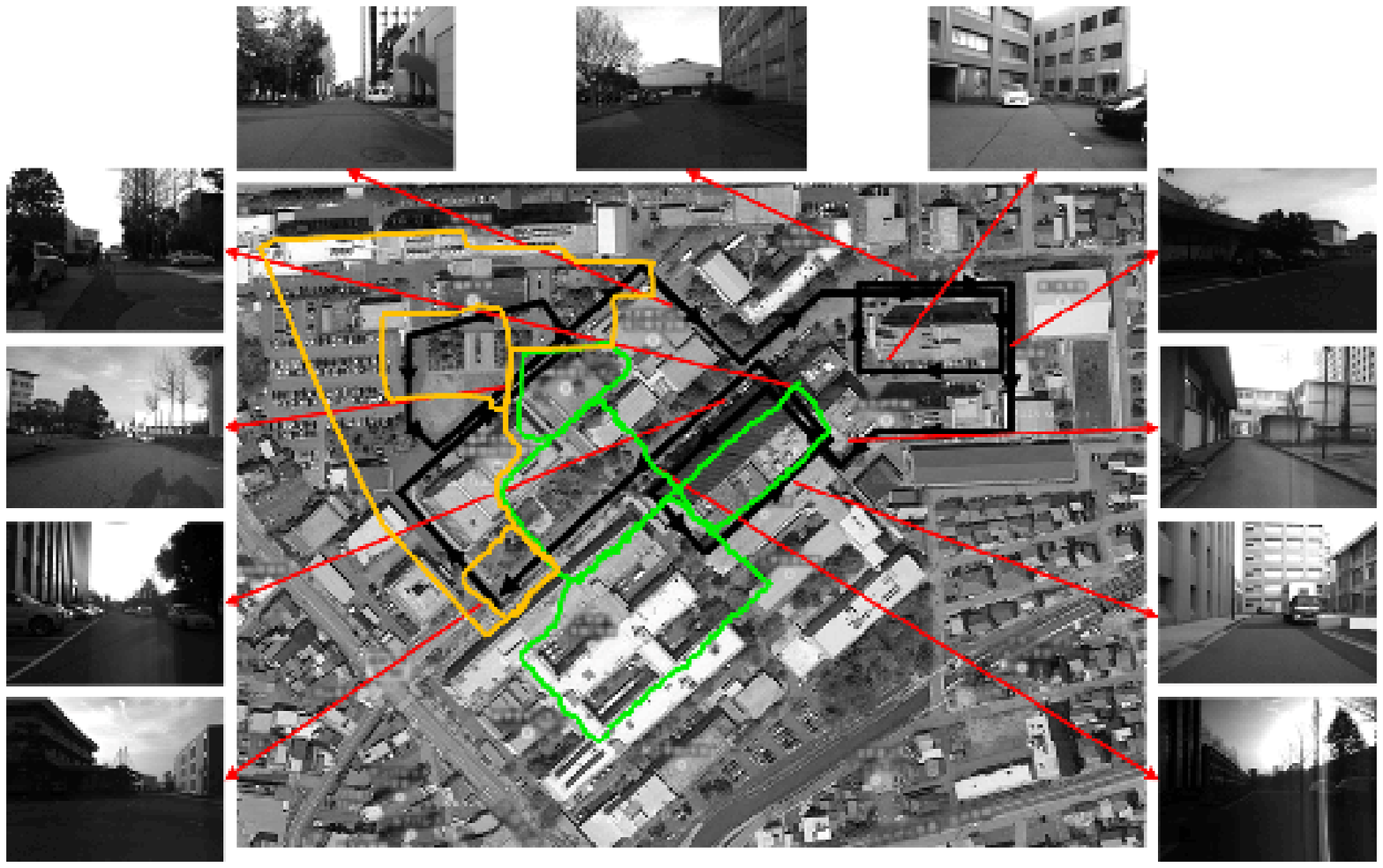}{a}\\
\FIG{4}{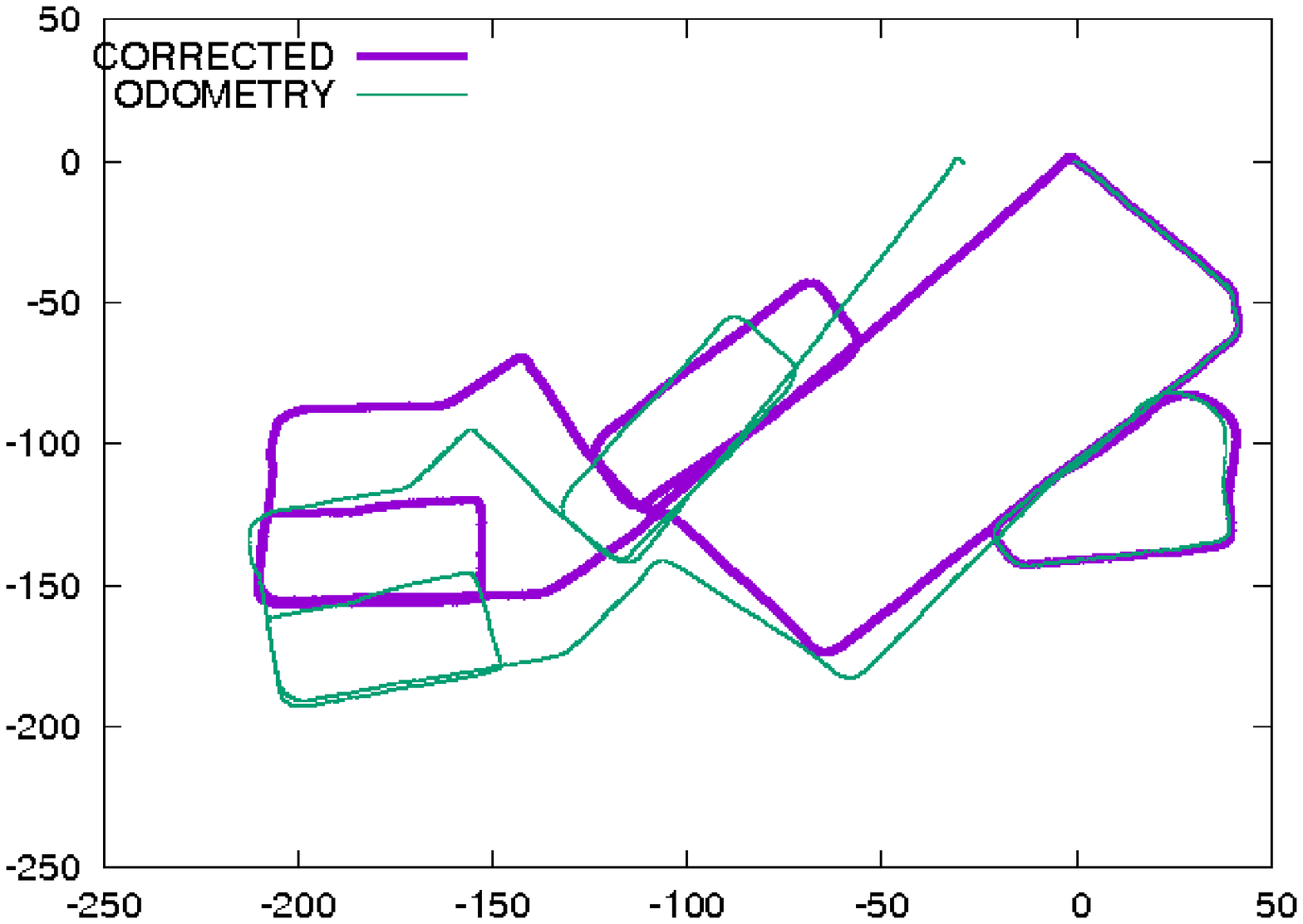}{b}%
\FIG{4}{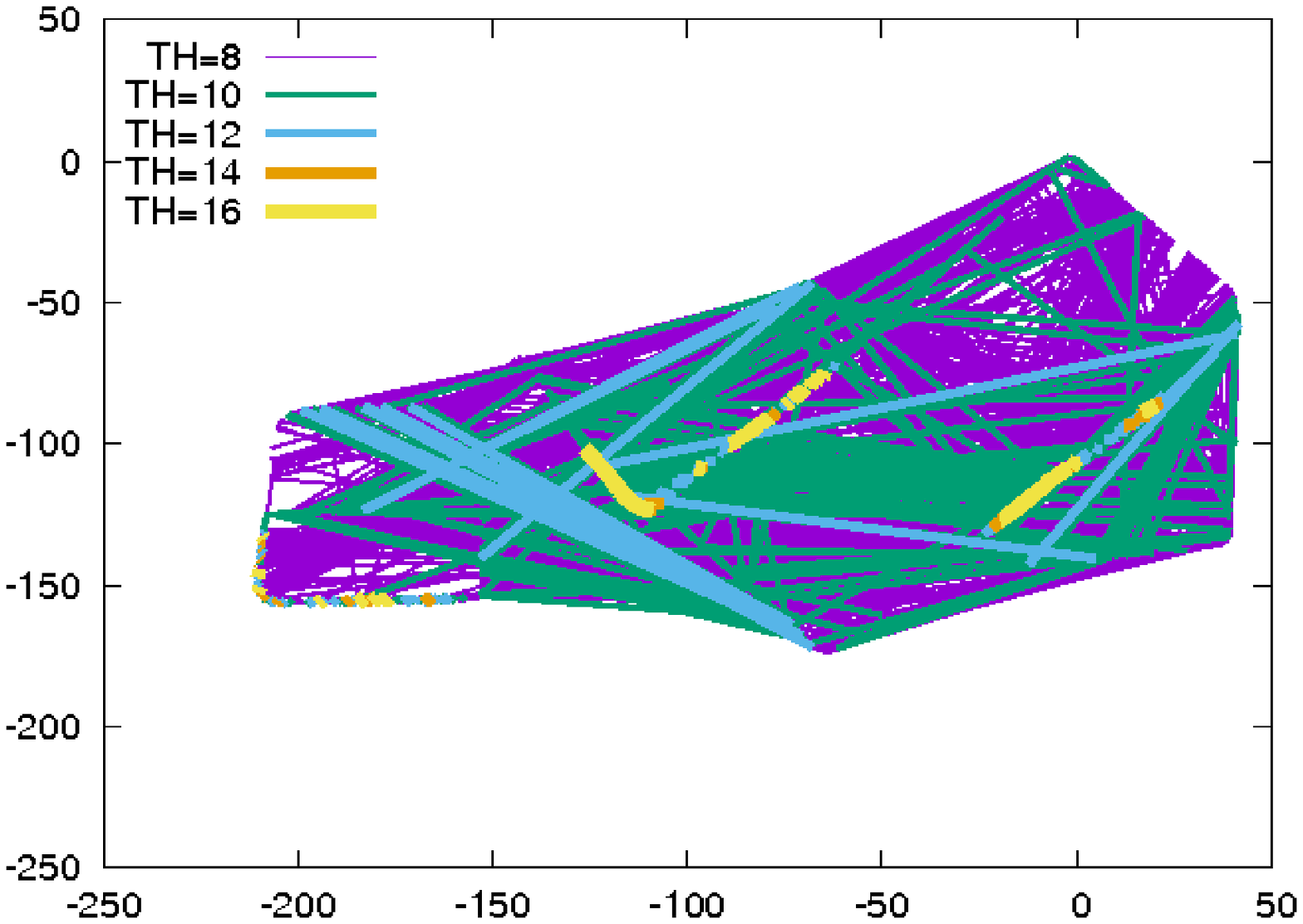}{c}\\
\FIG{4}{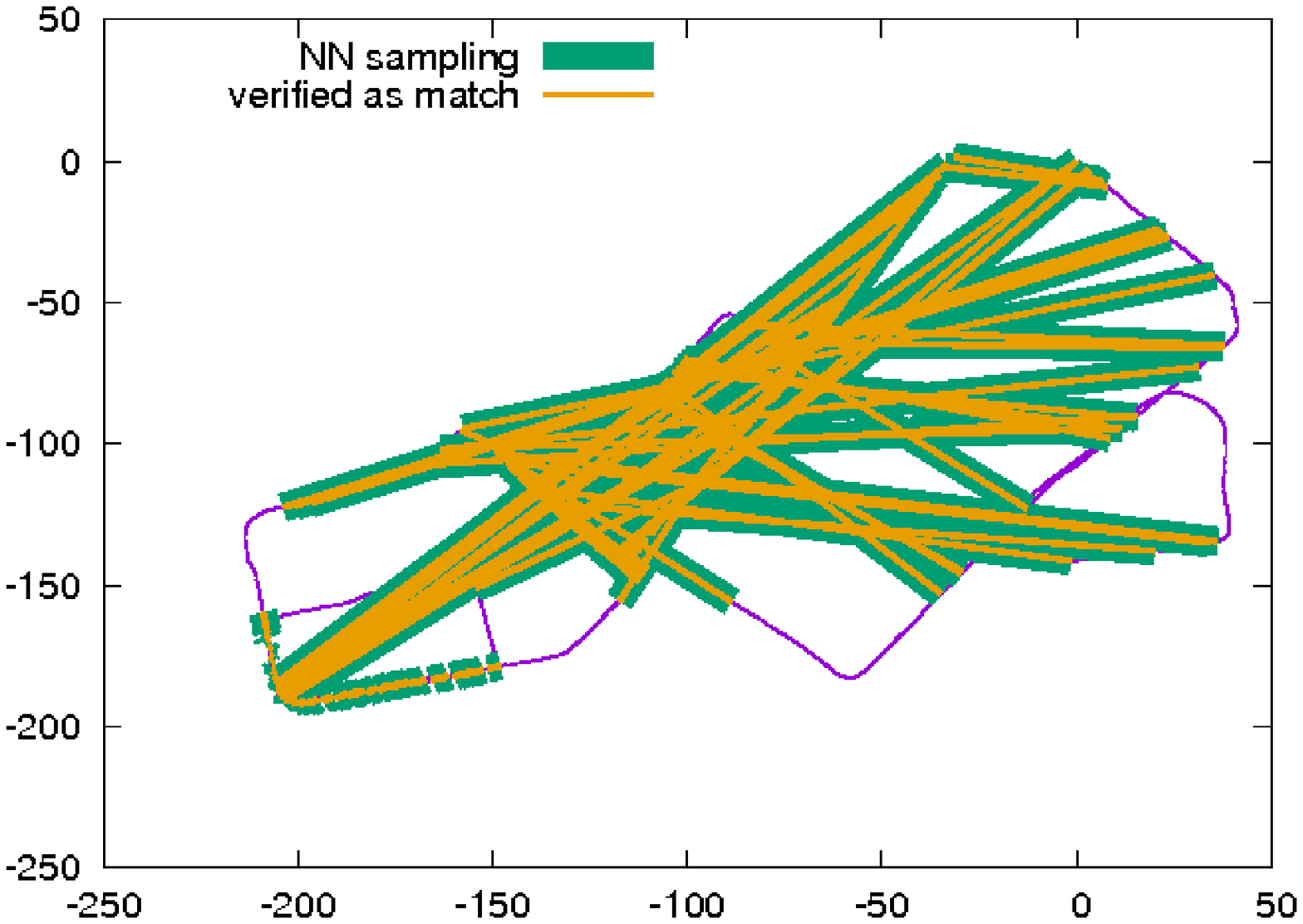}{d}%
\FIG{4}{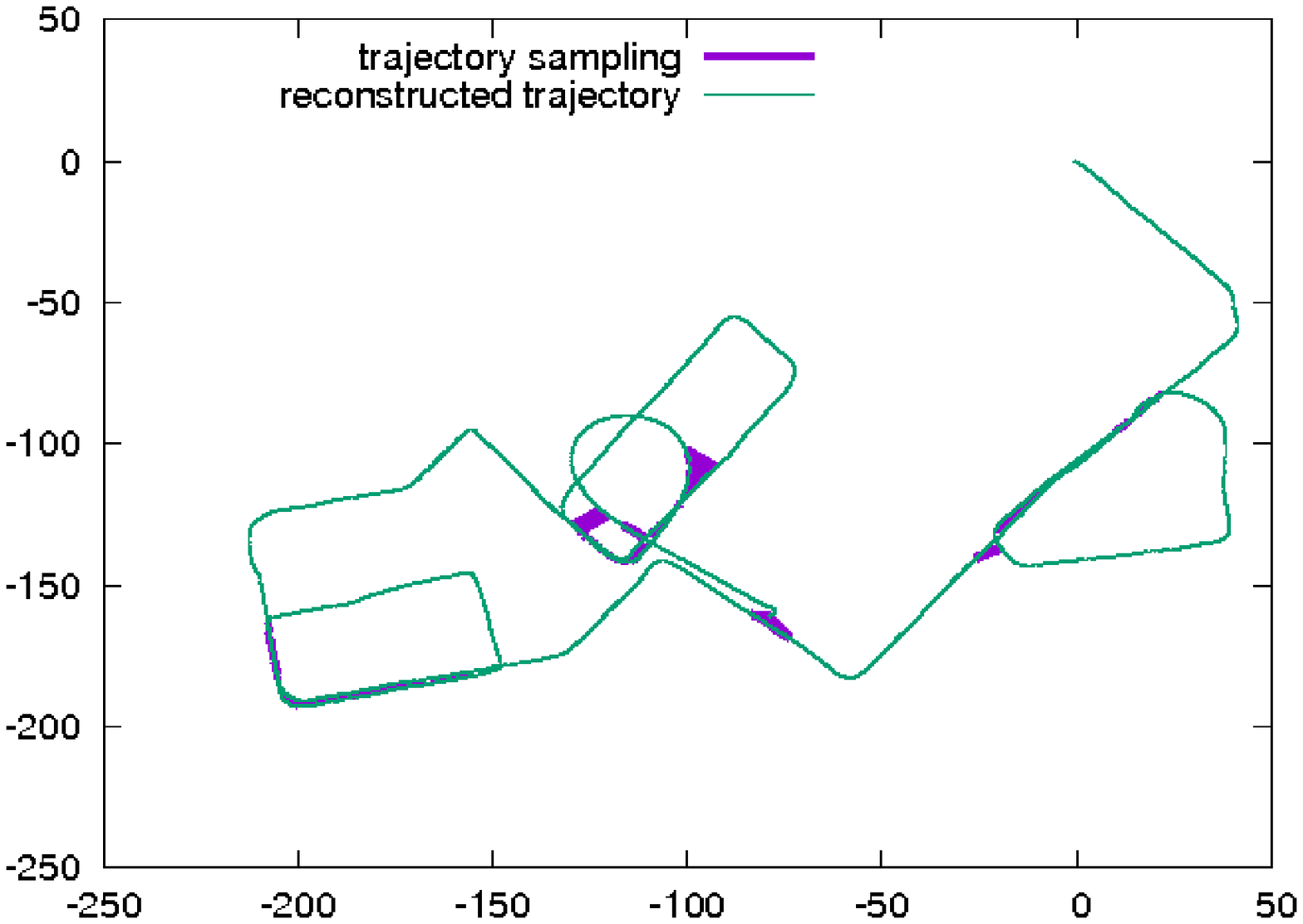}{e}\vspace*{-2mm}\\
\end{center}
\caption{Guided sampling for loop closure verification. Presented in (a) is our experimental environment with robot trajectory \#1 (black), \#2 (green), and \#3 (orange). Loop closure detection is an essential task for correcting the accumulated error in visual odometry (b). Indicated in (c) are typical results of detecting loop closure constraints by FAB-MAP image retrieval + RANSAC post-verification, for different settings of RANSAC thresholds, TH=8,10,12,14,16. One can see that the detection performance in terms of precision and recall is significantly less than perfect and that high recall is achieved when TH= 8 or 10 at the cost of extremely low precision. The basic idea of guided sampling is to verify loop closure hypotheses in a planned order (rather than in conventional uniform order) by exploiting a domain specific knowledge of mutual consistency between loop closure constraints. Illustrated in (d) and (e) are sample results of guided sampling using two different strategies, called NN sampling and trajectory sampling.}\label{fig:S}
\vspace*{-5mm}
\end{center}
\end{figure}
}

\newcommand{\figT}{
\begin{figure}[b]
\begin{center}
\begin{center}
\FIGR{8}{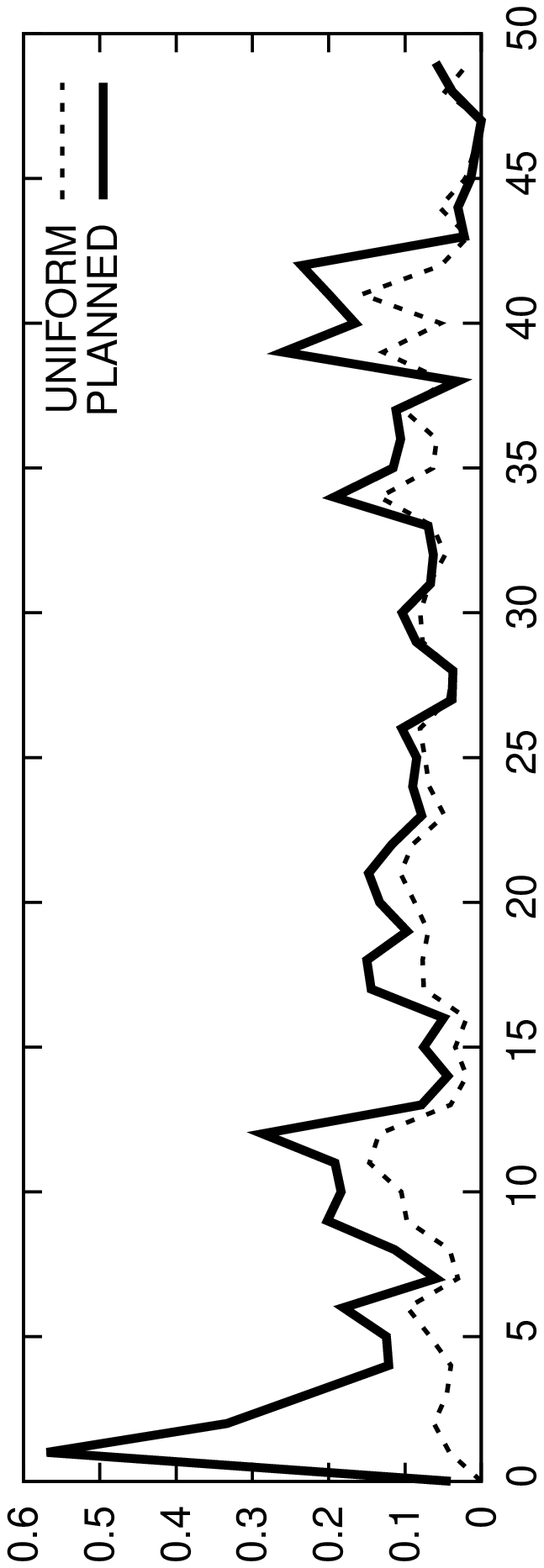}{a}
\FIGR{8}{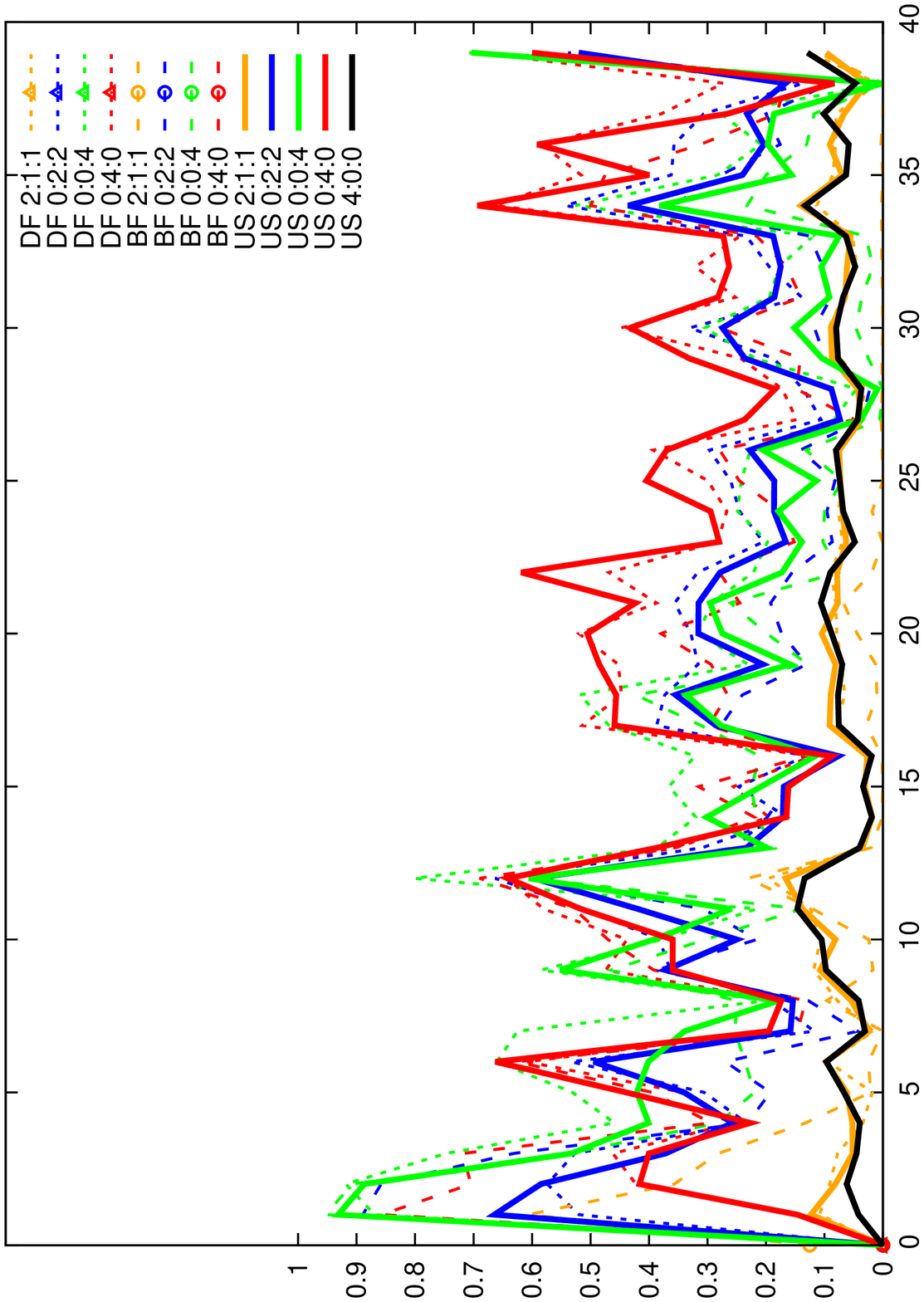}{b}
\end{center}
\caption{Effectiveness of using robot trajectory hypotheses as a guide for
sampling loop closure constraints. (a) is a histogram of the success ratio of
detecting correct loop closure hypotheses for two distinct cases: sampling
is guided (``PLANNED") and not guided (``UNIFORM"). The horizontal
axis represents the accuracy of the robot trajectory hypotheses, in terms of error
[m] in robot location on the reconstructed trajectory with respect to the
ground-truth trajectory. (b) compares success ratio for different combinations of guided sampling strategies. In the legend, ``DF", ``BF", and ``US"
are the depth-first, breadth-first, and uniform sampling strategies, respectively,
in Section \ref{sec:3b}; ``x:y:z" indicates that the parameters
$P_{TS}, P_{NS}$
for switching strategies are set to
$P_{TS}=z/(x+y+z)$
and
$P_{NS}=y/(x+y+z)$.
}\label{fig:T}
\vspace*{-5mm}
\end{center}
\end{figure}
}

\newcommand{\figUa}[2]{%
\begin{minipage}[b]{4cm}
\FIG{3.4}{#2.eps}{}\vspace*{-1mm}\\
\hspace*{-2mm}\FIGR{3.5}{#1_#2.eps}{}%
\end{minipage}%
\hspace*{-5mm}%
}

\newcommand{\figU}{
\begin{figure*}[t]
\begin{center}
\begin{center}
\figUa{neighbor}{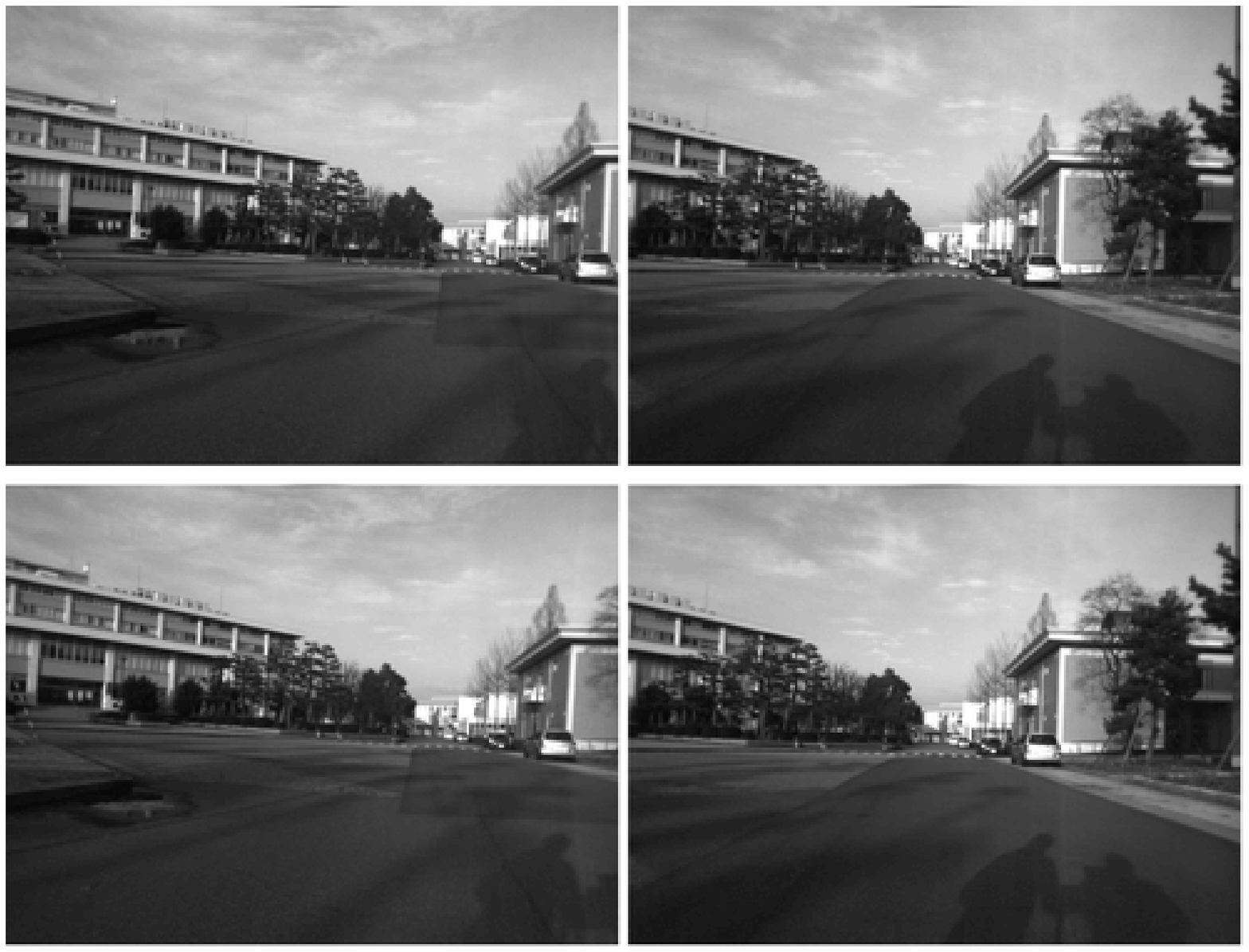}%
\figUa{neighbor}{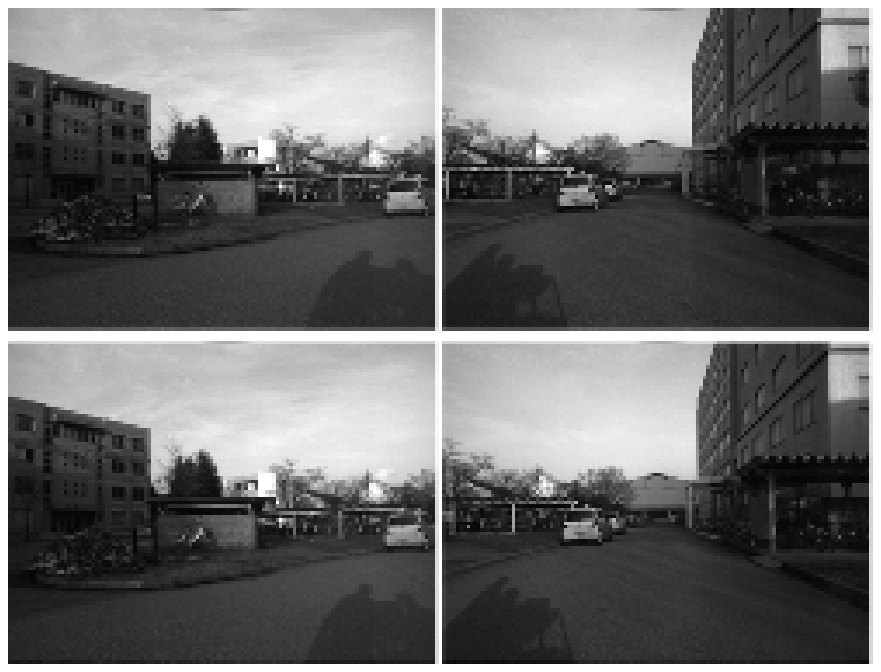}%
\figUa{neighbor}{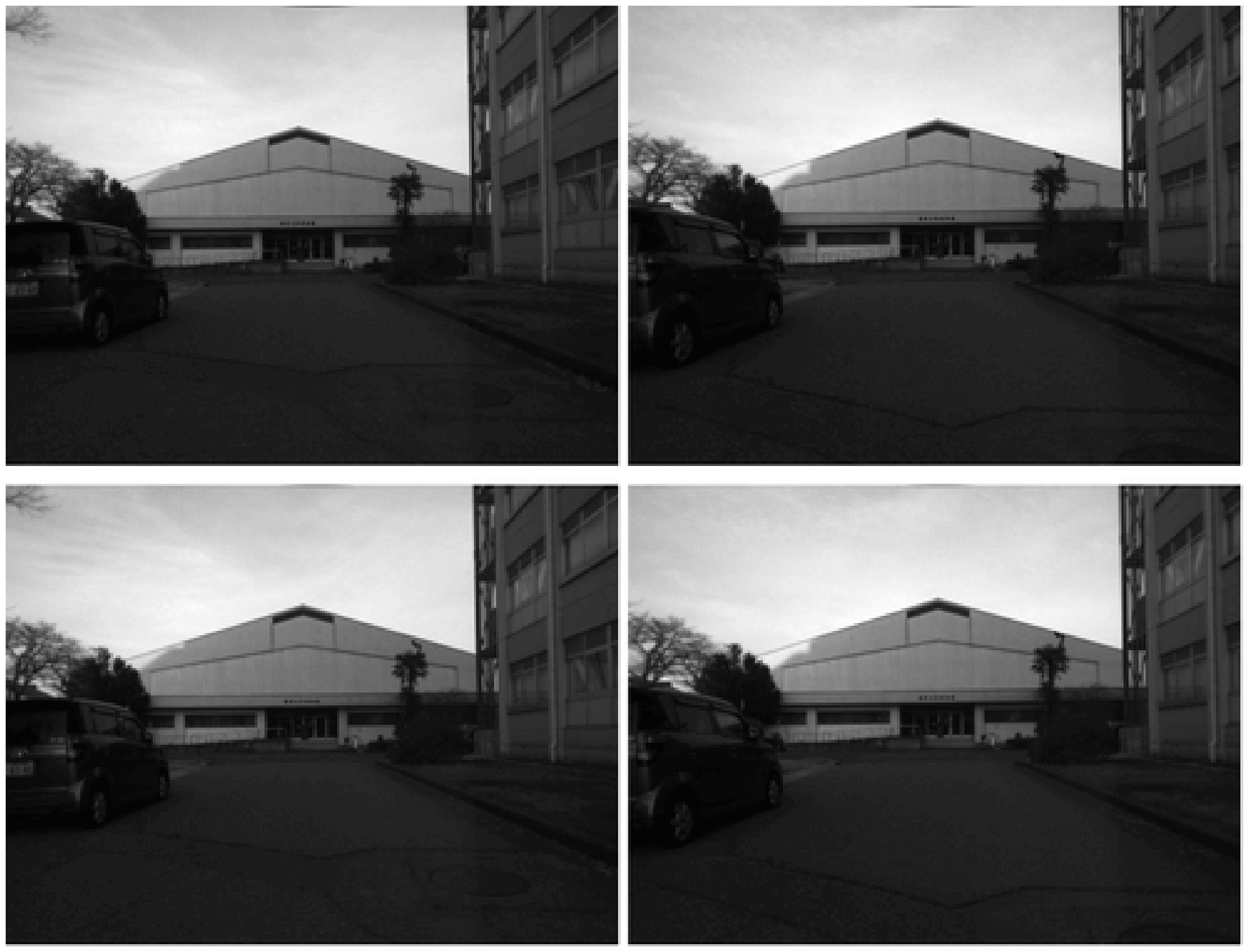}%
\figUa{neighbor}{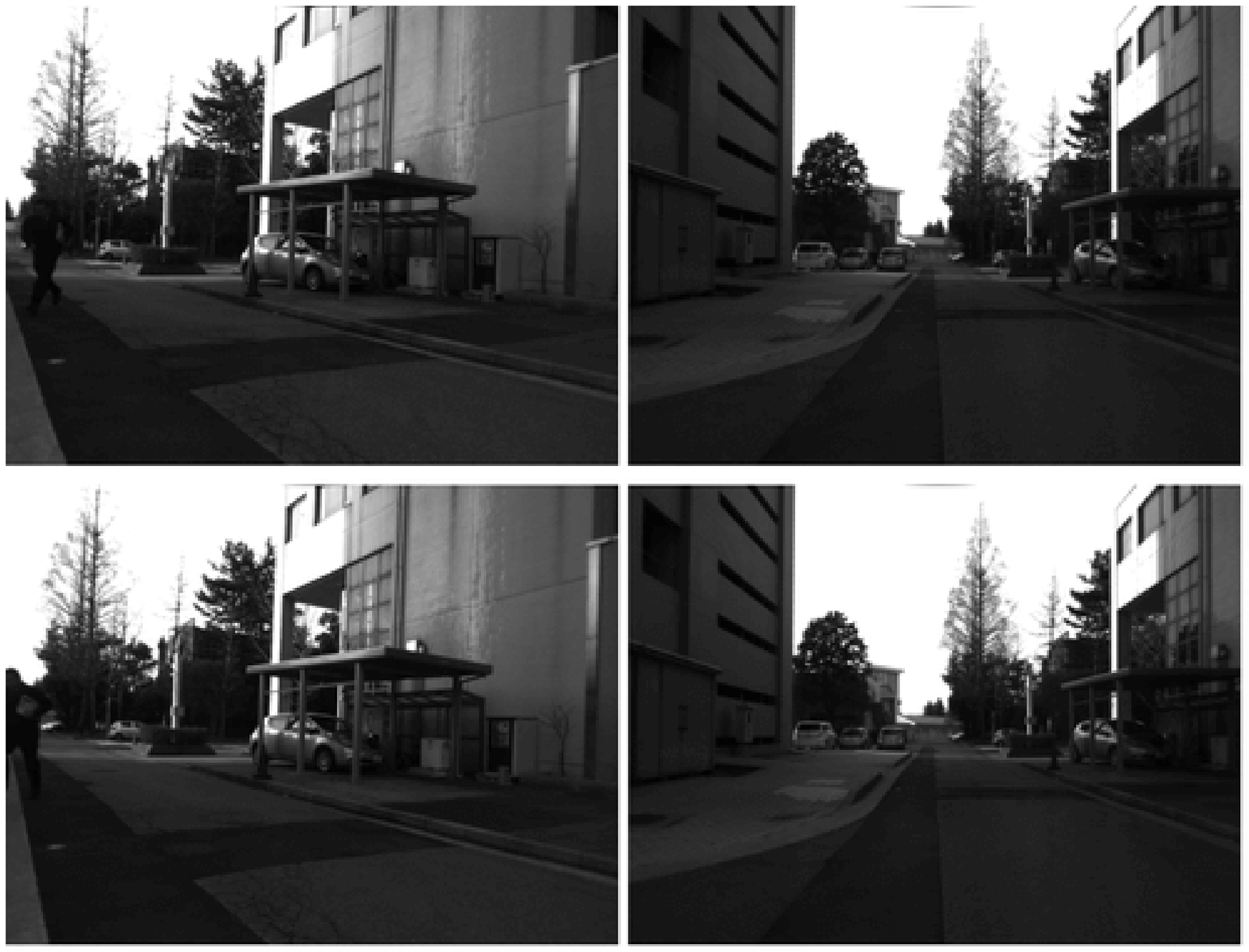}%
\figUa{neighbor}{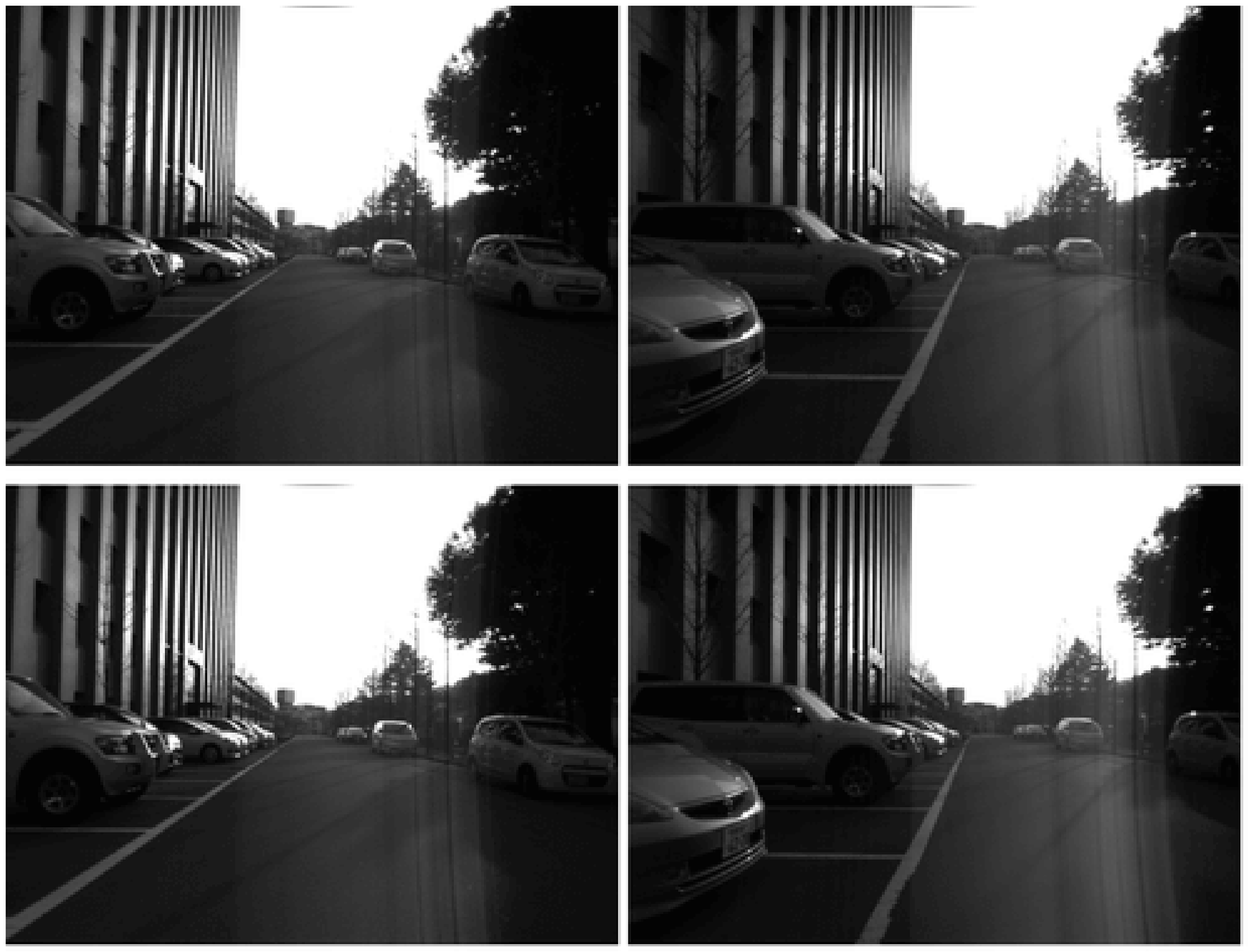}\vspace*{-2mm}\\
{\scriptsize neighbor sampling}\vspace*{2mm}\\
\figUa{trajectory}{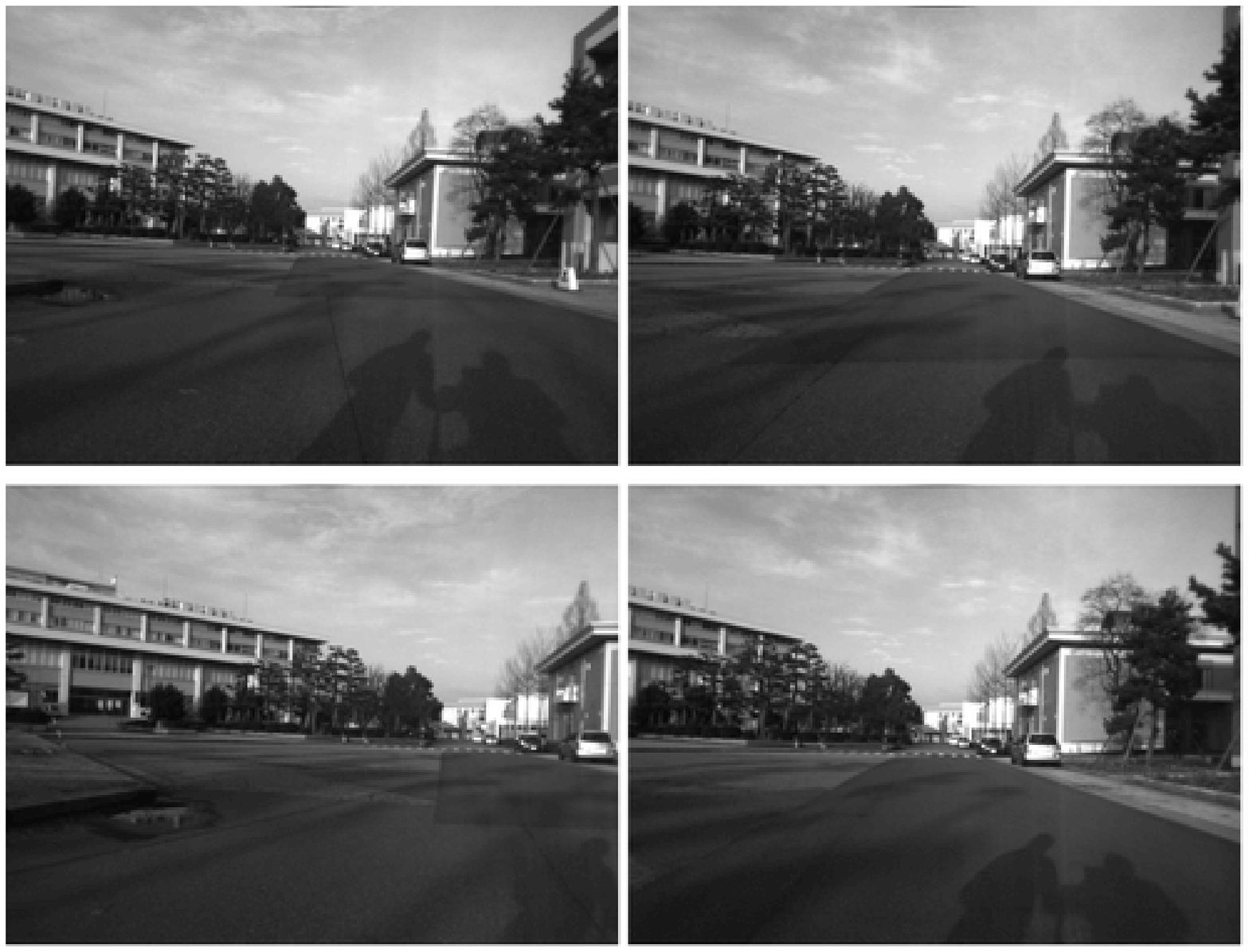}%
\figUa{trajectory}{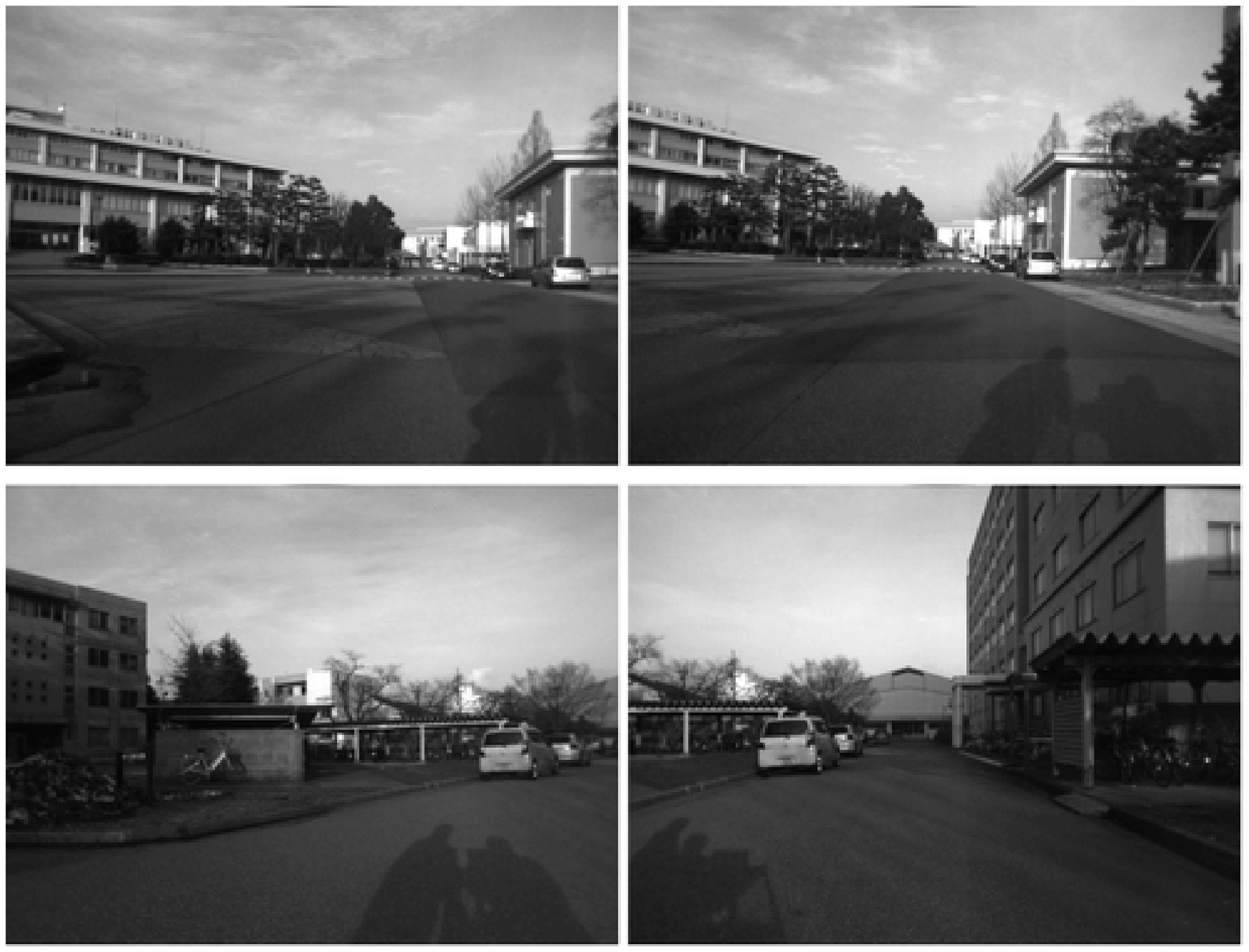}%
\figUa{trajectory}{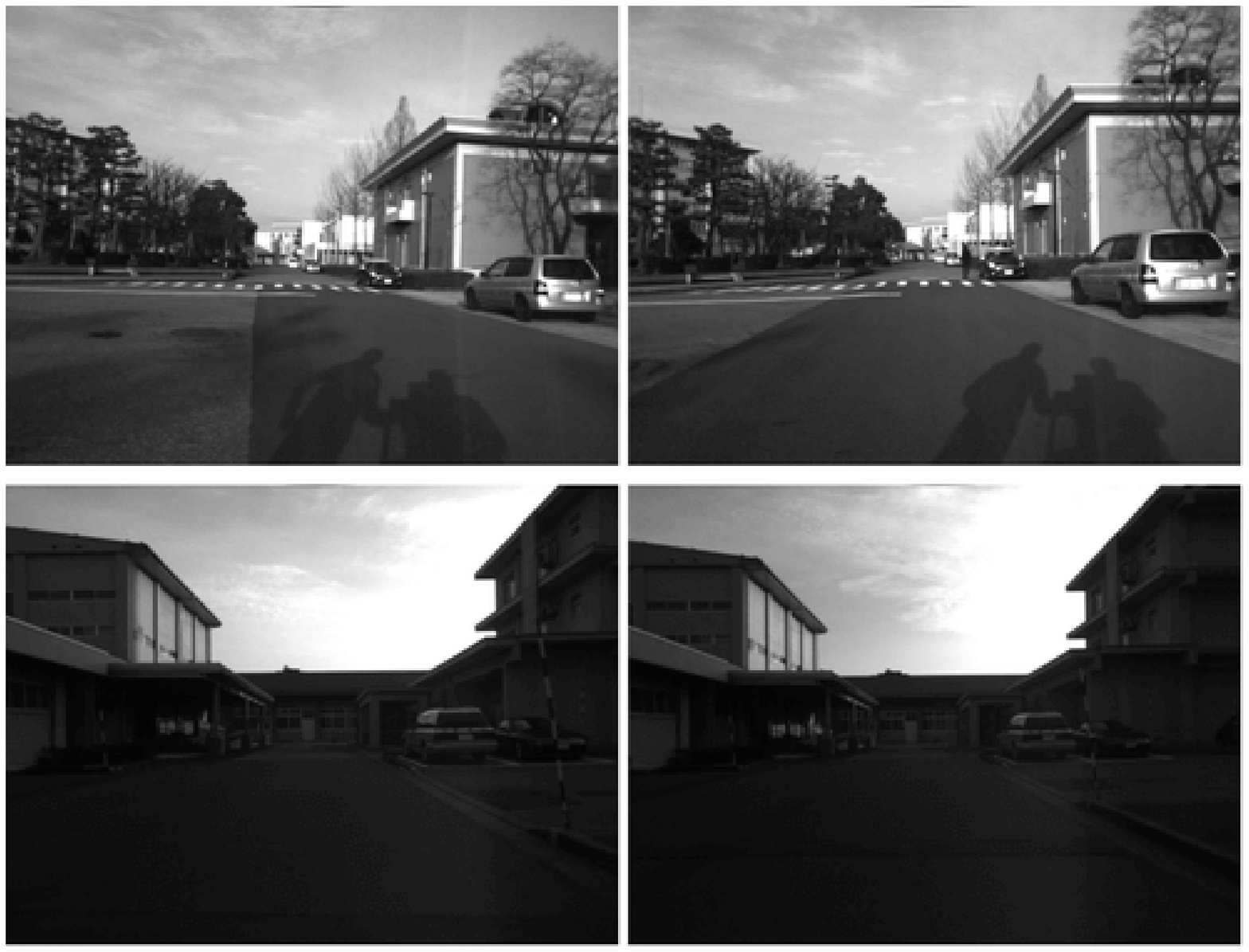}%
\figUa{trajectory}{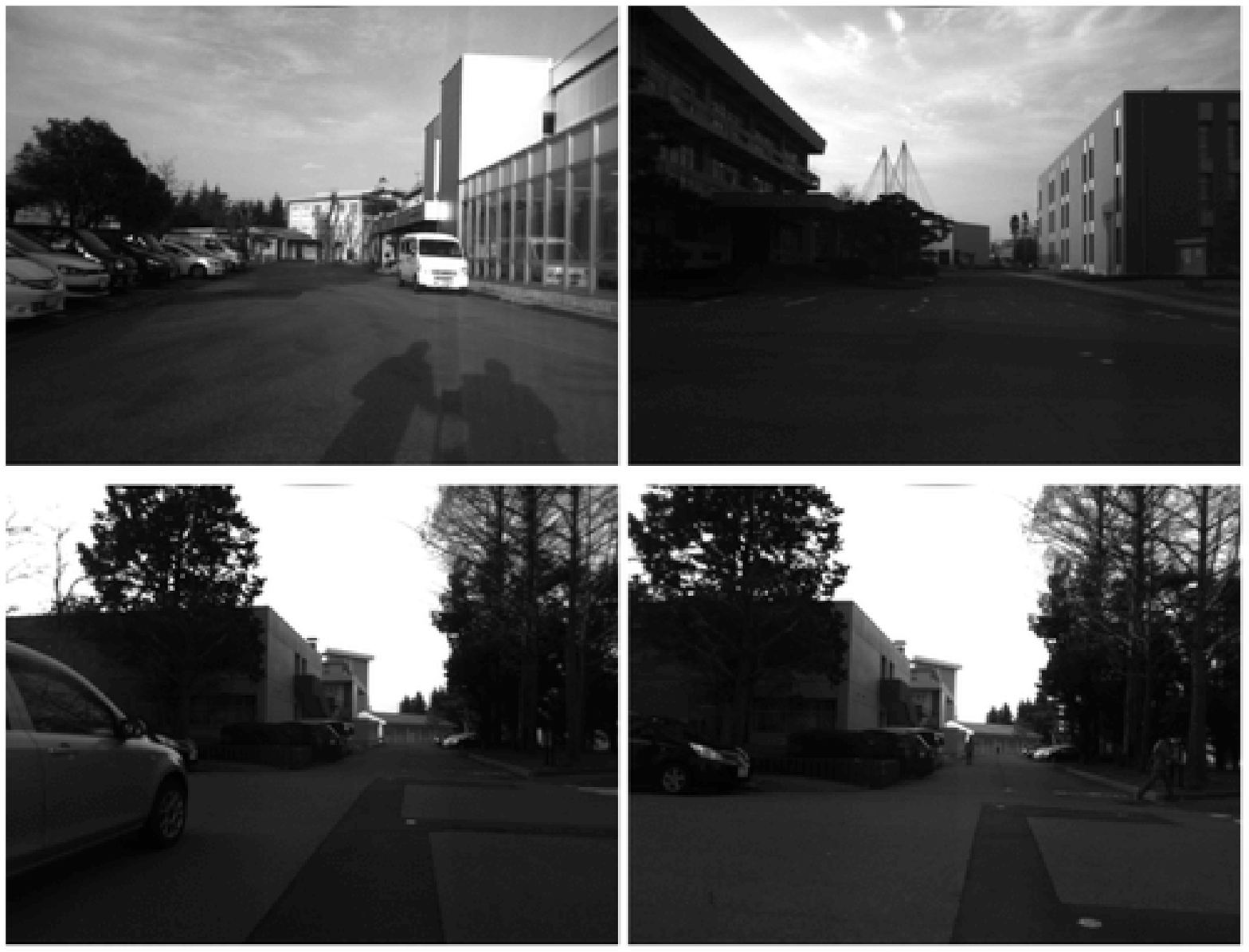}%
\figUa{trajectory}{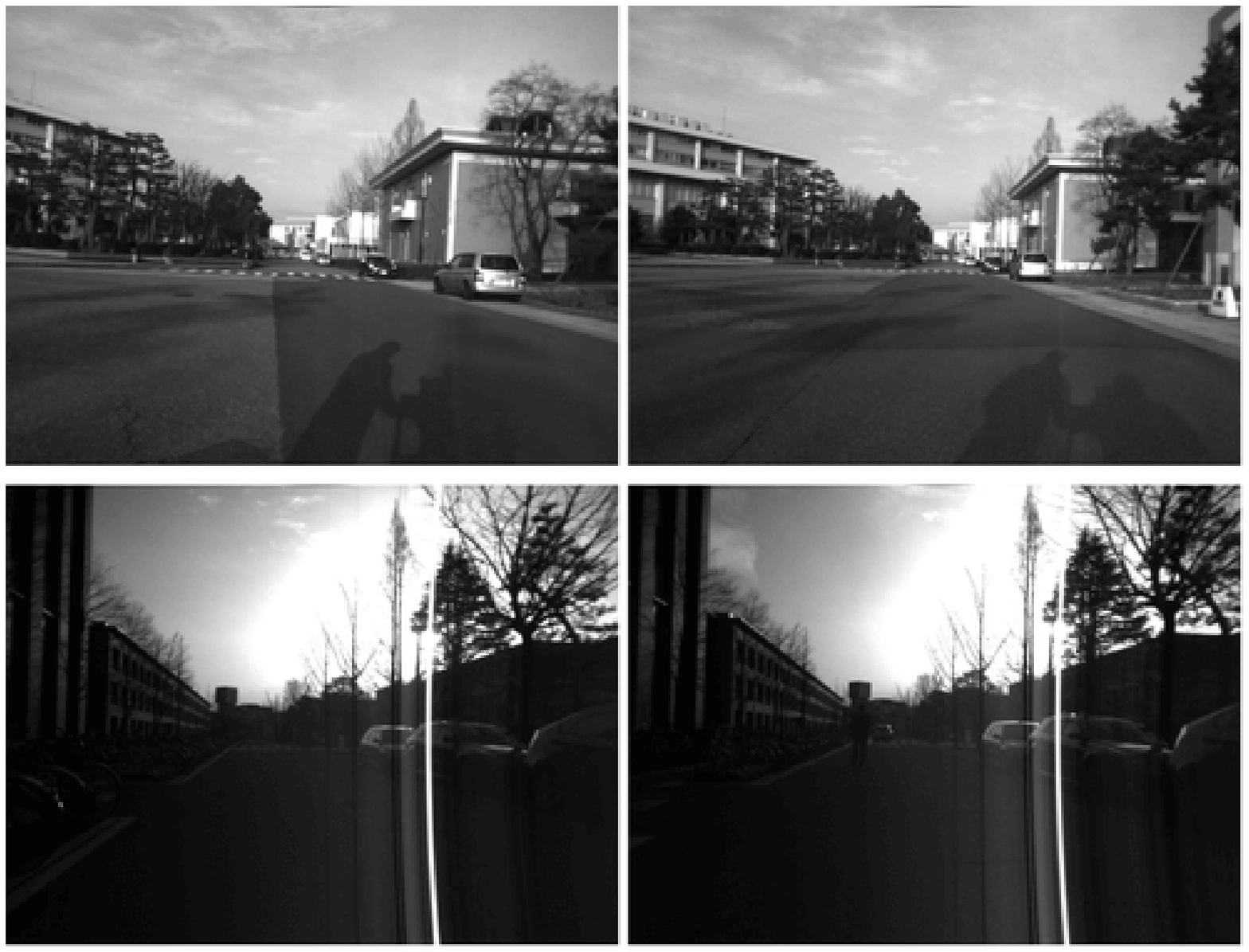}\vspace*{-2mm}\\
{\scriptsize trajectory sampling}\\
\end{center}
\caption{Examples of correct loop closure constraints proposed by two different guided sampling strategies. For each example, the top and bottom panel
represent two different loop closure constraints $s^{top}=(i,j)$ and $s^{bottom}=(i',j')$ 
using the corresponding image pairs
$(i,j)$ and $(i',j')$, 
where the former
constraint $s^{top}$ is used as a guide to sample the latter (and correct) constraint $s^{bottom}$. }\label{fig:U}
\vspace*{-5mm}
\end{center}
\end{figure*}
}

\newcommand{\figW}{
\begin{figure}[b]
\begin{center}
\FIG{4}{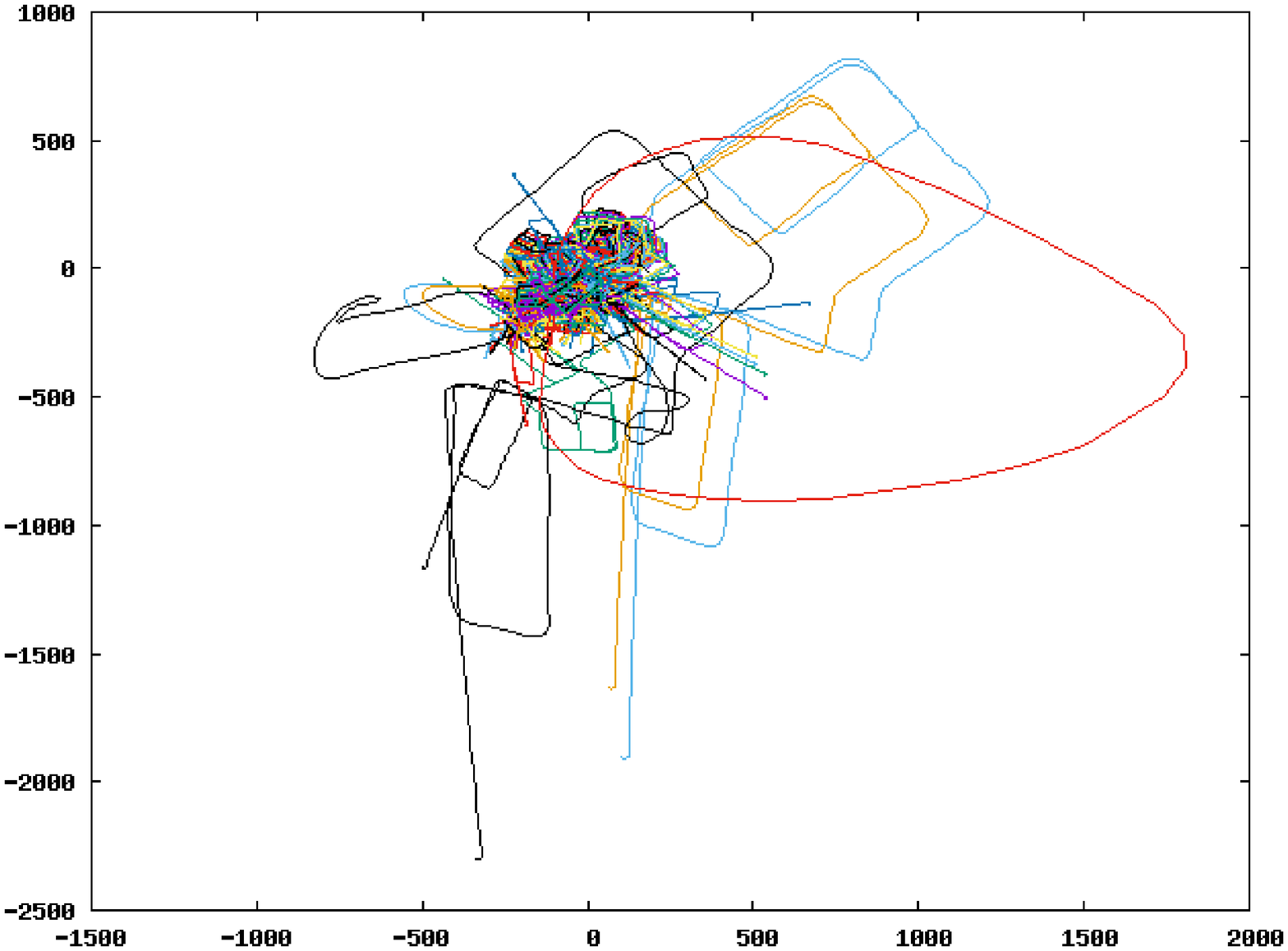}{}\hspace*{-3mm}%
\FIG{4}{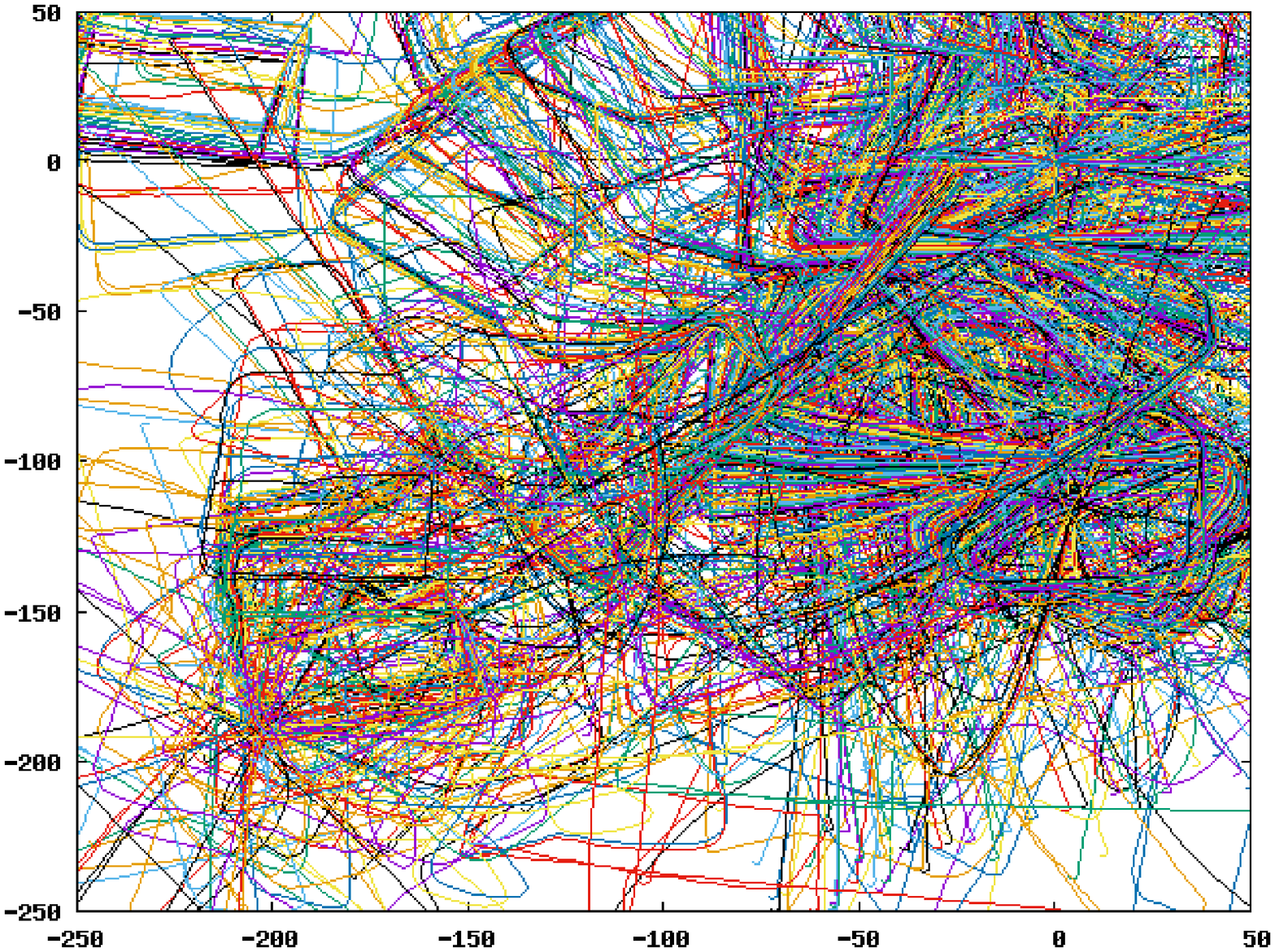}{}
\end{center}
\caption{All of the 538 robot trajectory hypotheses generated in Experiment \#1. On the right is a close-up of the figure on the left.}\label{fig:W}
\end{figure}
}

\title{\LARGE \bf
Incremental Loop Closure Verification by Guided Sampling 
} 


\maketitle

\begin{abstract}
Loop closure detection, the task of identifying locations revisited
by a robot in a sequence of odometry and perceptual observations,
is typically formulated as a combination of two subtasks: (1) bag-of-words image retrieval and (2) post-verification using RANSAC
geometric verification. The main contribution of this study is the
proposal of a novel post-verification framework that achieves good
precision recall trade-off in loop closure detection. This study is
motivated by the fact that not all loop closure hypotheses are
equally plausible (e.g., owing to mutual consistency between loop
closure constraints) and that if we have evidence that one hypothesis
is more plausible than the others, then it should be verified more frequently.
We demonstrate that the problem of loop closure detection can be viewed as an instance
of a multi-model hypothesize-and-verify framework and build guided sampling strategies on the framework where loop closures proposed
using image retrieval are verified in a planned order (rather than
in a conventional uniform order) to operate in a constant time.
Experimental results using a stereo SLAM system confirm that the proposed
strategy, the use of loop closure constraints and robot trajectory
hypotheses as a guide, achieves promising results despite the fact
that there exists a significant number of false positive constraints and
hypotheses.
\end{abstract}

\section{Introduction}\label{sec:intro}

Loop closure detection, the task of identifying locations revisited by
a robot in a sequence of odometry and perceptual observations, is an
important first step to estimate robot trajectory and has contributed to
important applications including 
localization \& mapping \cite{hahnel2003}, landmark discovery \cite{cpd13}, image alignment \cite{zhou2015flowweb}, 
topological mapping \cite{blanc2005indoor},
and place recognition \cite{lcd5}.
Failure in loop closure detection can yield catastrophic damage in
an estimated robot trajectory, and achieving an acceptable tradeoff between
precision and recall is critical in this context. In previous research 
\cite{lcd13,lcd6,lcd5},
loop closure detection is typically formulated as a bag-of-words image
retrieval problem where a query/database image is represented by an
unordered collection of vector quantized local invariant features termed
visual words and then efficiently indexed and retrieved to identify
pairs of matched locations using each view image as a query input.
However, even state-of-the-art image retrieval techniques generate a considerable number of false positives owing to confusing features and perceptual
aliasing \cite{lcd5}. 
Therefore, the image retrieval process is typically followed
by a post-verification step using robust RANSAC verification methods \cite{loransac}. 
However, the post-verification step is a computationally intensive
process, 
which requires quadratic time in the length of a view sequence. 
Therefore, improving the total cost for performance of loop
closure detection is an important practical problem and main focus of
this study.

\figS

In this study, we address the above issue with guided sampling. Unlike
previous frameworks where latest loop closure constraints $z_t$ (proposed
by image retrieval) are verified individually in a uniform order at each
location $t$, in the proposed framework, a constant number of constraints are
intelligently sampled in a planned order from all the $O(t^2)$ constraints
obtained to date. This study is motivated by the fact that not all loop
closure hypotheses are equally plausible (e.g., owing to mutual
consistency between loop closure constraints \cite{lcd26}) and that if we have
evidence that one hypothesis is more plausible than the others, then it should
be verified more frequently. Examples of such plausible hypotheses include:
\begin{enumerate}
\item
a loop closure constraint that provides a plausible reconstruction
of a robot trajectory;
\item
a loop closure constraint that is consistent with a plausible robot
trajectory hypothesis;
\item
a loop closure constraint that is spatially similar to a plausible loop
closure constraint.
\end{enumerate}
To implement the above idea, we cast loop closure detection as an
instance of a multi-model hypothesize-and-verify problem where a
set of hypotheses of robot trajectory is hypothesized from loop closure
constraints and verified in terms of consistency against other loop
closure constraints. The proposed approach is motivated by three independent
observations. First, we are inspired by the recent success of guided
sampling strategies in hypothesize-and-verify techniques (e.g., USAC \cite{raguram2013usac}). Second, loop closure detection is essentially a multi-model
estimation problem \cite{kanazawa2004detection}, 
rather than the single model estimation considered in classical applications of the hypothesize-and-verify approach
(e.g., structure-from-motion \cite{raguram2013usac}), 
where the goal is to identify multiple
instances of models (i.e., loop closure hypotheses) and the inliers to
one model behave as pseudo-outliers to the other models. Finally, and
most importantly, the framework is sufficiently general and effective
for implementing various guided sampling strategies that implement
the domain knowledge presented above.

Although the proposed approach is general, 
we focus on a challenging scenario of stereo SLAM,
which has been attracting increasing interest in recent years \cite{ppniv14dellaert},
to demonstrate the efficacy of the proposed system. Our experiments employ a stereo SLAM system that implements odometry
using stereo visual odometry as in \cite{geiger2011stereoscan}, loop closure detection using
appearance-based image retrieval with SURF local features and bag-of-words image model as in \cite{lcd5}, post-verification using RANSAC
geometric verification on local feature keypoints as in \cite{loransac}, and pose graph SLAM as in \cite{kaess2008isam}. Fig.\ref{fig:S}b,c illustrate an odometry-based robot trajectory, trajectory corrected by loop closing, and a set of loop closure
constraints determined by employing the two strategies.
As can be observed, significant errors in trajectory are accumulated as the robot navigates
and the errors are successfully corrected given correct loop closure constraints. It
can also be seen that image retrieval-based loop closure detection
and RANSAC post-verification are both less than perfect (Fig. \ref{fig:S}c);
there are a significant number of false positives and negatives. These two
types of errors, accumulated errors in odometry and misrecognition in detection and post-verification of errors, are the main
error sources that we address in this study. Experimental results
using our stereo SLAM system confirm that the proposed strategy, the use of loop
closure constraints and robot trajectory hypotheses as a guide, achieves
promising results despite the fact that there are a significant number of false
positive constraints and hypotheses (Fig. \ref{fig:W}).

\subsection{Related Work}

This study can be viewed as a novel application of multi-model-based consensus approaches (e.g., multi-RANSAC \cite{kanazawa2004detection}). Similar to previous multi-model approaches, we focus on determining multiple plausible hypotheses (i.e., loop closure hypotheses) rather than the single best hypothesis.

We allow a set of detected models (i.e., loop closure constraints) to be
partially inconsistent with each other. Such partial inconsistency in loop
closure constraints can be resolved reliably by employing modern
SLAM back-ends such as robust pose graph optimization in \cite{lcd26}.

Most of the existing works on loop closure detection have focused on
the image retrieval step in the task, rather than the post-verification step
\cite{lcd2}. In fact, loop closure detection techniques are typically classified
in terms of image retrieval strategies (rather than post-verification
strategies) \cite{lcd3}. Images are typically represented by a collection of
invariant local descriptors \cite{lcd5} or a global holistic descriptor \cite{lcd6,lcd9}. Loop closure detection has been employed by many SLAM
systems \cite{lcd19,lcd7,lcd14,lcd15}. However, the above works did not focus on
the post-verification step or introduce novel insight to the guided
sampling strategy.

Guided sampling has been studied in many matching problems
that are closely related to loop closure detection with respect to its
objective. In \cite{lcd27}, techniques were presented for improving the speed of robust
motion estimation based on the guided sampling of image features, which
is inspired by the MLESAC algorithm. \cite{lcd10} presented an approach called
double-window optimization where a place recognition is used to
identify loop-closing constraints and incorporate the constraints into the
optimization. \cite{lcd22} explored an approach based on considering features
in the scene database and matching them to query image features
as opposed to previous methods that match image features to visual
words or database features, and presented an efficient solution based
on prioritized feature matching. \cite{lcd23} proposed a method for improved
geometric verification by exploiting the statistics of image collections and
gathering information during geometric verification, to improve the
overall efficiency. \cite{lcd25} presented an unsupervised learning approach
for learning threshold for geometric verification. Recently, a series of
robust SLAM back-end algorithms that allow misrecognition in loop
closure constraints have been studied \cite{lcd26}. However, their objectives
are neither the guided sampling nor SLAM front-end applications;
rather, they follow conventional batch-style matching. Moreover,
differing from general purpose matching algorithms, we are interested
in and focus on the use of task specific knowledge regarding loop closure
detection to improve overall performance.

This paper is a part of our studies on loop closure detection. Recently, we
have discussed cross-season place recognition \cite{iros15a}, part-based scene
modeling \cite{icra15b}, landmark discovery \cite{icra15a}, and map descriptor \cite{ppniv15} in IROS15, ICRA15, PPNIV15 papers. Guided sampling in loop closure detection has not been addressed in the above papers.

\section{Approach}

\subsection{Loop Closure Detection}\label{sec:3a}

For clarity of presentation, we first describe a baseline SLAM
system where the proposed approach is built and
used as a benchmark for performance comparison in the experimental
section. As mentioned, we build the proposed system on a stereo SLAM
system where a stereo vision sensor is employed for both 
visual odometry \cite{geiger2011stereoscan} and visual feature acquisition \cite{lcd5} and follow the
standard formulation of pose graph SLAM \cite{grisetti2010tutorial}. 
In pose graph SLAM,
the robot is assumed to move in an unknown environment, along a
trajectory described by a sequence of random variables 
$x_{1:T}=$ $x_1$, $\cdots$, $x_T$.
While moving, it acquires sequences of odometry measurements
$u_{1:T}=$ $u_1$, $\cdots$, $u_T$
and perception measurements
$z_{1:T}=$ $z_1$, $\cdots$, $z_T$. 
Each odometry measurement
$u_t$ $(1\le t\le T)$
is a pairing of rotation
and translation acquired by visual odometry. Each
perceptual measurement
$z_t$
is a set of loop closure constraints
$z_t^1$, $\cdots$, $z_t^{N_t}$,
each of which is a pair of location IDs,
$t$, $t'$
with a likelihood
score that represents the likelihood of the location pair belonging to
the same place, which is acquired by FAB-MAP in our case. More
formally, we begin with an empty list of loop closure constraints.
At each time $t$,
we execute FAB-MAP using the latest visual image as
a query to identify the top 
$N_t=50$
ranked images that receive the highest
likelihood scores. We then insert the
$N_t=50$
pairs from the query image and each of the
$N_t$
top-ranked database images as new constraints to the list.

For simplicity, we begin by assuming that fixed sets of loop closure constraints
$z_{1:T}$ and map hypotheses $m_{1:M}$ 
are a priori given; typical hypothesize-and-verify algorithms require such a fixed set assumption \cite{kanazawa2004detection}. 
Clearly, this assumption is violated in our SLAM applications as both the loop closure constraints and the map
hypotheses must be incrementally derived as the robot navigates. This 
incremental setting will be addressed in Section
\ref{sec:3c} 
by relaxing
the fixed set assumption. We divide the entire measurement sequence
into constant time windows and generate one hypothesis per window.
To generate a hypothesis, we employ pose graph SLAM that expects the following as
input: (1) the single loop closure constraint whose score
received from FAB-MAP is the highest within the time window of interest,
and (2) a sequence of previous odometry measurements. This yields
$M=T/W$
map hypotheses when the size of the time window is $W$.
In experiments,
we set the time window size sufficiently small,
$W=10$,
that appearance of images do not change significantly within time window.

Performance of loop closure detection is typically evaluated by
precision-recall. This performance measure requires a set of ground
truth loop closure constraints and a set of constraints verified as matched
by the RANSAC. 
We run the RANSAC for 
each loop closure constraint 
(selected by the guided sampling)
that consists of a pair of images 
to check if the keypoint configuration is geometrically consistent between the image pair.
We use RANSAC
verification with the fundamental matrix in \cite{loransac}
and a preset threshold of ``8",
as indicated in Fig.\ref{fig:S}c.
If the RANSAC score exceeds the threshold, the input pair is verified as
matched. For each query image
$i$,
we define a range of ground-truth loop closure constraints in the form:
$(i,j^{begin}_i)$, 
$\cdots$,
$(i,j^{end}_i)$, 
and consider a
verified constraint
$(i,j)$
is correct if and only if
$j\in [j^{begin}_i, j^{end}_i]$.

Based on the above terminology, we formulate the problem of
guided sampling in loop closure detection. Let $s$ denote a selection
of a loop closure constraint. Recalling that a set of $N_t$ new constraints
arrive at each time instance $t$, we represent a selection $s$ of loop closure
constraints by:
\begin{equation}
s = ( t, n ) ~~~ t \in [1,T], n \in [1, N_{t}].
\end{equation}
Let $v_i$ denote the result of post-verification:
\begin{equation}
v_i = V( s_i ) ~~~~ v_i \in \{0,1\}, 
\end{equation}
which indicates if the RANSAC score exceeds a predefined threshold (``1") or not (``0"). Guided sampling is the problem of selecting the next constraint to verify:
\begin{equation}
s_i = S ( u_{1:t}, z_{1:t}, s_{1:i-1}, v_{1:i-1} ),
\end{equation}
given a history of previous odometry $u_{1:t}$ and perception $z_{1:t}$, and a history of previous selections $s_{1:i-1}$, and verification results $v_{1:i-1}$.

\figW

\subsection{Guided Sampling Strategies}\label{sec:3b}

A naive strategy for guided sampling is to uniformly sample
one constraint/hypothesis from a history of previously acquired
constraints/hypotheses. This strategy is straightforward and easy to implement. Unfortunately, it does not achieve acceptable precision-recall
tradeoff as indicated in the experimental section (Section \ref{sec:4}). To achieve
improved precision-recall tradeoff, we present several different strategies
for guided sampling in the following discussion.

The first strategy, termed {\bf trajectory sampling (TS)}, is a strategy
that samples loop closure constraints that are consistent with a robot
trajectory hypothesis (Fig. \ref{fig:S}e). The basic idea is to verify the consistency
between a pair of constraints using the robot trajectory hypothesis as
an intermediate. This strategy selects one random robot trajectory
hypothesis $h$ and then samples a loop closure constraint, a pair of IDs
$(t,t')$
of locations that are near each other:
\begin{equation}
|| p(t,h) - p(t',h) || < T_p, \label{eqn:2}
\end{equation}
where $p(t,h)$ is the 2-dimensional coordinate of location $t$ conditioned on a robot trajectory hypothesis $h$, and $T_p$ is a preset threshold, 10 m.

The second strategy, termed {\bf neighbor sampling (NS)}, is a strategy
that samples loop closure constraints that are neighbors to plausible
constraints. 
Fig. \ref{fig:S}d
presents examples of neighbor sampling. This strategy is motivated by the fact that in street-like environments, there often
exists a sequence of matched views, rather than single isolated matches.
This strategy selects one random trajectory hypothesis and 
one random verified loop closure constraint $(i,j)$
on the selected
trajectory, and then samples one of its four neighbors
$(i\pm 1, j\pm 1)$.

The third strategy, termed {\bf breadth first (BF)}, is a strategy that
samples trajectory hypotheses in a breadth-first order, rather than in a
uniform order. This strategy is motivated by a limitation of uniform
samplings in an incremental scenario, that is, the total number of old sampling hypotheses generated at the beginning of a robot's navigation tends
to be considerably greater than the new hypotheses generated at the end of
the navigation. As a result, the distribution of ``being sampled" becomes
considerably unbalanced among the hypotheses as indicated in Fig. \ref{fig:P}b ``UNIFORM SAMPLING". To address this limitation, this strategy continuously
monitors the number of individual hypotheses being sampled to date and
selects the hypothesis with the least number of being sampled as the next
hypothesis to sample.

The fourth strategy, termed {\bf depth first (DF)}, is a strategy that
samples a trajectory hypothesis according to the 
{\it importance weight}, which
is evaluated by the number of loop closure constraints 
that 
are consistent with the hypothesis of interest
and have been
verified as matched to date. Currently, we divide the hypothesis set into
two subsets, upper half and lower half, according to the importance
weight, and then sample the next hypothesis from the former subset.

We need to select one guided sampling strategy at a time among the several
different strategies for the sampling constraints (TS, NS) and hypotheses
(BF, WS) discussed above. Currently, we introduce a random selection
rule using preset parameters 
$P_{TS}$, $P_{NS}$, $P_{BF}$ and $P_{DF}$. We randomly
select one of the three different strategies for hypothesis sampling, BF, DF,
and uniform strategies with probabilities
$P_{BF}$ : $P_{DF}$ : ($1-$$P_{DF}-$$P_{BF}$).
We also randomly select one of the three different strategies for constraint
sampling, TS, NS, and uniform strategies with probabilities
$P_{TS}$ : $P_{NS}$ : ($1-$$P_{TS}-$$P_{NS}$).
In experiments, we test the different settings of
these probabilities to investigate the contributions of the different sampling
strategies.

\CO{
\subsection{Guided Sampling Strategies}

{\scriptsize
\begin{tabular}{llllll}
PARAM & BF & DF & uniform & NS & TS \\
1 & n & n & 1.0 & 0.0 & 0.0 \\
2 & n & n & 0.0 & 1.0 & 0.0 \\
3 & n & n & 0.0 & 0.0 & 1.0 \\
4 & n & n & 0.0 & 0.5 & 0.5 \\
5 & n & n & 0.5 & 0.25 & 0.25 \\
6 & y & n & 0.0 & 1.0 & 0.0 \\
7 & y & n & 0.0 & 0.0 & 1.0 \\
8 & y & n & 0.0 & 0.5 & 0.5 \\
9 & y & n & 0.5 & 0.25 & 0.25 \\
10 & n & y & 0.0 & 1.0 & 0.0 \\
11 & n & y & 0.0 & 0.0 & 1.0 \\
12 & n & y & 0.0 & 0.5 & 0.5 \\
13 & n & y & 0.5 & 0.25 & 0.25 \\
\end{tabular}
}

}

\subsection{Incremental Extension}\label{sec:3c}

In this section, we relax the fixed set assumption in Section \ref{sec:3a} and
consider the general incremental setting of loop closure detection.
As a main extension, the system must update the set of loop
closure constraints and set of map hypotheses. Further, because
sampling strategies introduced in Section \ref{sec:3a} rely on the knowledge
of mutual consistency between loop closure constraints and map
hypotheses, it also must update this knowledge every time a new
constraint or hypothesis arrives. Thus, we introduce the concept
of {\bf consistency matrix}: $C$,
which is an $N_tT \times M$ sized matrix
defined as consistency $C_{ij}$ between each $i$-th
loop closure constraint and $j$-th map hypothesis, 
and we incrementally update it every time a new
constraint or hypothesis arrives. The update for a new constraint and 
hypothesis requires $M$ and $N_tT$ computations
of consistency, respectively. In the moderate-sized environments 
(e.g., $M\le 1000$, $N_tT\le 1\times 10^5$) 
considered in the experimental section, we can assume these two
costs are negligible compared to the costs for loop closure
verification and map reconstruction.

\figT

\figP

\section{Experiments}\label{sec:4}

\subsection{Settings}

We conducted loop closure detection experiments using a stereo SLAM system on a university campus. Our experiments employed a stereo SLAM
system that implemented the proposed guided sampling strategies.
The principal steps involved visual odometry, loop closure detection, and post-verification. The first step executed stereo visual odometry to
reconstruct the robot trajectory. We adopted the stereo visual odometry
algorithm proposed in \cite{geiger2011stereoscan}, which has proven to be effective in recent visual odometry
applications (e.g., \cite{brubaker2013lost}). The second step applied the appearance-based
image retrieval, FAB-MAP, proposed in \cite{lcd5}. This step generated a set
of $N_t$ new loop closure constraints and inserted these into the constraint
list as indicated in Section \ref{sec:3a}. The third step performed guided sampling
to select a set of loop closure constraints to verify and applied each of
these to RANSAC verification to be classified as matched or unmatched.

Fig. \ref{fig:S}a,b present the ground-truth robot trajectories and visual odometry
trajectories superimposed on Google map imagery. The
ground-truth trajectories were generated using a SLAM algorithm based
on the graph optimization in \cite{kaess2008isam} using manually identified ground-truth
loop closure constraints as input. As indicated, significant odometry errors
were accumulated as the robot navigated. We collected three sequences
along routes with travel distances of 1364 m, 1020 m, and 1250 m, using
a cart equipped with a Bumblebee stereo vision camera system, as
illustrated in Fig.\ref{fig:S}a. We defined a ground-truth loop closure constraint
as a pairing of two locations $i,j$ whose distance was less than 10 m.
Occlusion was severe in the scenes and people and vehicles were
dynamic entities occupying the scenes. We processed each path and collected
stereo image sequences with lengths 5759, 6358, and 6034.

\figU

\subsection{Proof-of-Concept Experiment}

In this paper, we rely on an assumption that robot trajectory hypotheses act as a guide to sample a new good loop closure constraint. One could argue that because an excessive number of false positive hypotheses exist, as indicated in Fig. \ref{fig:W}, that the guide could cause more harm than assistance. As a proof-of-concept experiment, we experimentally compared the
ratio of successful loop closure detection for three distinct cases: (1)
not guided, (2) guided by a correct robot trajectory hypothesis, and
(3) guided by an incorrect robot trajectory hypothesis. To evaluate (2)
and (3), we collected sets of robot trajectory hypotheses with different
levels of correctness, which were measured in terms of percentage of
correct loop closure constraints hypothesized by the hypothesis of
interest. Fig. \ref{fig:T} presents the success ratio for loop closure detection guided
by robot trajectory hypotheses. It can be seen that the proposed guided
sampling does not harm the process, even when an inaccurate (false positive)
hypothesis is used and that it does help when an accurate hypothesis is used.
We can observe that sampling guided by false positive hypotheses frequently
behaves similarly to uniform sampling, as the robot's poses on false positive
hypotheses typically distribute uniformly over the environment, as
indicated in Fig. \ref{fig:W}.

\subsection{Precision-Recall Performance}

We evaluated the proposed strategies for guided sampling. We employed precision and recall as introduced in Section \ref{sec:3a} as the performance
measure. 
We did not use trivial loop closure constraints 
where the travel distance between the image pairs was overly short, 
less than 100 m.
That is, loop closures were not detected 
until the robot had travelled 100 m from the start
location in all experiments.

Fig.\ref{fig:P}a presents the precision-recall curves. It can be observed that the TS strategy method outperformed all the other strategies considered in the
current experiments. We observe that TS strategy was able to use robot trajectory hypotheses as an effective guide to propose good loop closure constraint, despite the fact that majority of robot trajectories are partially incorrect with respect to the ground truth as shown in Fig. \ref{fig:W}.

\subsection{Examples of Guided Sampling}

Fig. \ref{fig:U} presents examples of correct loop closure constraints proposed
by guided sampling. We tested two different strategies: neighbor
sampling, which uses a previously verified loop closure constraint as a guide and samples its neighbor constraints, and trajectory sampling, which uses robot trajectories reconstructed from previously verified loop closure constraints as a guide and samples constraints that are consistent with a reconstructed trajectory. It can be seen that the neighbor sampling strategy contributed to detecting loop closure constraints of similar scenes to previously detected loop closure constraints; whereas, the trajectory
sampling strategy tended to detect loop closure constraints of extremely dissimilar scenes.

\subsection{Ratio of Verified/Matching Constraints}

Fig. \ref{fig:P}c and Fig. \ref{fig:P}d present the ratio of loop closure constraints that are verified
as matched and are correct with respect to the ground-truth. It
can be seen that the neighbor sampling strategy outperformed the other strategies
considered in finding matched and correct constraints.
We observe that 
neighbor sampling strategy was effective to sample good constraints especially when almost all the robot trajectories reconstructed to date are incorrect with respect to the ground-truth.

\section{Conclusions}

The main contribution of this study is the
proposal of a novel post-verification framework that achieves good
precision recall trade-off in loop closure detection. 
We showed the loop closure detection can be viewed as an instance of multi-model hypothesize-and-verify framework, 
and based on an incremental extension of this framework,
we built strategies for guided sampling,
by which loop closures proposed by image retrieval are verified in a planned order rather than in a conventional 
uniform order to operate in a constant time.
Experimental results using a stereo SLAM system confirmed that the proposed
strategy, the use of loop closure constraints and robot trajectory
hypotheses as a guide, achieves promising results despite the fact
that there exists a significant number of false positive constraints and
hypotheses.

\bibliographystyle{IEEEtran}
\bibliography{lcd}

\end{document}